\documentclass[11pt]{article}

\usepackage{acl}

\usepackage{times}
\usepackage{latexsym}
\usepackage{xurl}
\usepackage[T1]{fontenc}
\usepackage[utf8]{inputenc}
\usepackage[most]{tcolorbox}
\usepackage{microtype}
\usepackage{enumitem}
\usepackage{inconsolata}

\usepackage{booktabs}
\usepackage{multirow}
\usepackage{makecell}
\usepackage{amssymb} 

\usepackage{graphicx}
\usepackage{xcolor}
\usepackage{colortbl}
\usepackage{arydshln} 
\usepackage{amsmath}
\usepackage{pifont}
\usepackage{microtype}

\definecolor{overallbg}{RGB}{225, 213, 231}  
\definecolor{gainbg}{RGB}{213, 232, 212}     
\definecolor{gaintext}{RGB}{0, 100, 0}       

\definecolor{plusColor}{RGB}{200, 0, 0}      
\definecolor{minusColor}{RGB}{0, 0, 180}     
\definecolor{generalbg}{RGB}{235, 245, 255}  
\definecolor{medicalbg}{RGB}{255, 245, 235}  
\definecolor{graybg}{gray}{0.95}             

\definecolor{basegray}{gray}{0.92}           
\definecolor{methodblue}{RGB}{235, 245, 255} 

\newcommand{\p}[2]{#1\textsubscript{\textcolor{plusColor}{\textbf{+#2}}}} 
\newcommand{\m}[2]{#1\textsubscript{\textcolor{minusColor}{\textbf{#2}}}}  

\title{Thinking with Deltas: Incentivizing Reinforcement Learning \\ via Differential Visual Reasoning Policy}

\author{
    \textbf{Shujian Gao}\textsuperscript{1}, 
    \textbf{Yuan Wang}\textsuperscript{2}, 
    \textbf{Jiangtao Yan}\textsuperscript{3}, 
    \textbf{Zuxuan Wu}\textsuperscript{1, $\dagger$}, 
    \textbf{Yu-Gang Jiang}\textsuperscript{1, $\dagger$} \\
    \textsuperscript{1}Fudan University \quad
    \textsuperscript{2}Zhejiang University \quad
    \textsuperscript{3}Wuhan University
}

\begin{document}
\maketitle
{
  \renewcommand{\thefootnote}{\fnsymbol{footnote}} 
  \footnotetext[2]{Corresponding authors.} 
}
\begin{abstract}
Reinforcement Learning with Verifiable Rewards (RLVR) has significantly advanced reasoning capabilities in Large Language Models. 
However, adapting RLVR to multimodal domains suffers from a critical \textit{perception-reasoning decoupling}. 
Existing paradigms, driven by text-centric outcome rewards, reasoning in language medium, inadvertently encourage models to bypass visual perception. 
We empirically validate this through blind experiments: state-of-the-art policies maintain or surprisingly improve performance even when visual inputs are entirely removed. 
This reveals that these models degenerate into \textit{blind reasoners}, exploiting linguistic priors to generate plausible answers instead of attending to visual evidence.
In response, we propose \textbf{Thinking with Deltas}, a framework driven by a \textbf{Differential Visual Reasoning Policy (DVRP)}. 
DVRP introduces intrinsic supervision via visual triplets, comprising original, masked, and perturbed inputs. 
It optimizes the model to maximize reasoning divergence from masked inputs (enforcing \textit{visual sensitivity}) while minimizing divergence from perturbed inputs (ensuring \textit{visual robustness}). 
By aligning reasoning variations strictly with the \textit{Delta} of visual information, DVRP inherently bolsters visual understanding capabilities and significantly outperforms state-of-the-art methods on both general and medical benchmarks, without requiring external annotations or auxiliary tools.
\end{abstract}

\section{Introduction}
Recent advancements in Large Language Models (LLMs)~\cite{chang2024survey, brown2020languagemodelsfewshotlearners} have been substantially driven by Chain-of-Thought (CoT) prompting and the curation of high-quality reasoning data~\cite{kojima2022large, ye2025limo}.
Building on this foundation, Reinforcement Learning with Verifiable Rewards (RLVR)~\cite{ouyang2022traininglanguagemodelsfollow, shao2024deepseekmathpushinglimitsmathematical} has emerged as a promising post-training paradigm.
By incentivizing self-correction and scaling test-time compute, RLVR enables models to generate rigorous and self-consistent solution paths, establishing a new standard for complex problem-solving in textual domains \cite{wang2025reinforcementlearningreasoninglarge}.

However, a fundamental misalignment arises when directly applying the RLVR paradigm to multimodal domains~\cite{noisyrollout}.
Predominant approaches typically mirror text-centric methodologies ~\cite{deepseekr1} and rely heavily on textual outcome rewards such as accuracy or format constraints~\cite{wang2025beyond}.
Furthermore, attempts to incorporate process supervision via external judges often inherit the biases and limitations of the judge models themselves~\cite{luo2025ursa}.
Conceptually, if we analogize a Multimodal LLM (MLLM) to a human observer, the visual encoder functions as the eyes.
Optimizing rewards based solely on the output text sequence is equivalent to evaluating speech without verifying vision.
This approach neither ensures the effective perception of visual signals nor stimulates deep visual understanding~\cite{han2025learningseeingdemystifyingllm}.
Consequently, current frameworks neglect the causal dependency between visual inputs and reasoning outcomes.

\begin{figure*}[h]
  \includegraphics[width=\linewidth]{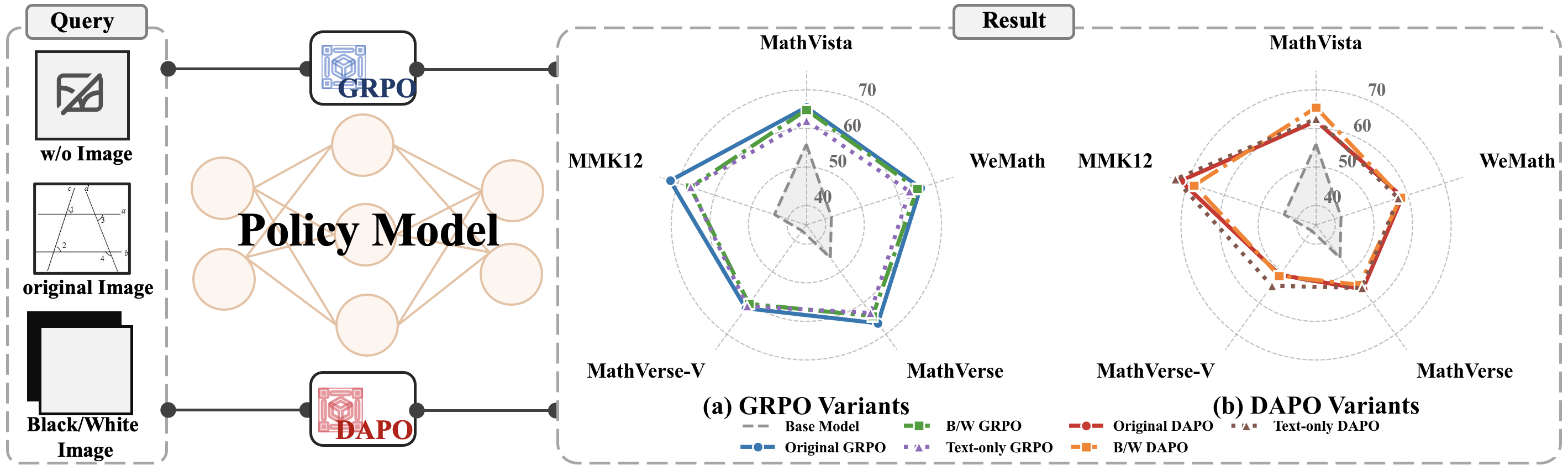}
    \caption{\textbf{Empirical Validation of Visual Decoupling.} 
    Performance comparison in blind settings (Text Only or Blank Image) reveals latent reward hacking. 
    The negligible performance drop in GRPO and the \textit{unexpected performance gain} in DAPO (where removing visual inputs actually improves accuracy) indicate that policies degenerate into \textit{blind reasoners} relying on linguistic shortcuts rather than visual evidence.}
  \label{fig:teaser}
\end{figure*}

As illustrated in Figure~\ref{fig:teaser}, we empirically validate this decoupling through \textbf{blind experiments} where visual inputs are either removed (Text Only) or replaced with blank images (Black/White).
We investigate the behaviors of DAPO and GRPO under varying input configurations.
For the GRPO baseline~\cite{shao2024deepseekmathpushinglimitsmathematical}, the observed performance drop is negligible.
This result is counter-intuitive, as the absence of essential visual contexts should theoretically precipitate a substantial performance collapse.
Even more strikingly, DAPO~\cite{dapo} exhibits an \textit{unexpected performance gain}: removing visual information \textit{improves} accuracy (e.g., \textbf{+3.5\%} on MathVista).
These findings suggest distinct \textit{reward hacking}, where the policy degenerates into a \textit{blind reasoner}.
Rather than grounding reasoning in visual perception, the model exploits linguistic shortcuts to maximize rewards, effectively treating visual data as distractive noise rather than indispensable evidence (Detailed analysis in Section~\ref{sec:rlvr_visual_dependency}).

To address this perception-reasoning decoupling without resorting to complex data construction~\cite{liu2024visual} or external reward engineering~\cite{zhang2025medtvtr1multimodalllmempowering}, we propose \textbf{Thinking with Deltas}, a novel framework centered around a \textbf{Differential Visual Reasoning Policy (DVRP)}.
Rather than relying on proxy textual rewards, DVRP introduces intrinsic visual supervision by constructing a visual triplet input stream: (1) \textit{Invariant} (the original image), (2) \textit{Decremental} (masked visual input), and (3) \textit{Incremental} (perturbed visual input).
Our core insight is that genuine visual reasoning must be strictly sensitive to the \textit{Delta} ($\Delta$) of visual information.
Specifically, DVRP enforces a differential policy that maximizes the divergence between the \textit{Invariant} and \textit{Decremental} states (proving visual sensitivity) while minimizing the divergence between the \textit{Invariant} and \textit{Incremental} states (ensuring visual robustness).
By optimizing these differential signals end-to-end, DVRP effectively compels the model to attend to visual evidence without requiring expensive external tools or dense annotations.

Our contributions are summarized as follows:
\begin{itemize}
    \item We identify the perception-reasoning decoupling in current multimodal RL as a primary bottleneck, arguing that standard objectives optimize for linguistic plausibility rather than visual grounding.
    
    \item We introduce the \textbf{DVRP}, an algorithm that leverages visual triplets to construct intrinsic \textit{Delta supervision signals} for both visual sensitivity and robustness, inherently bolstering visual perception and reasoning capabilities.
    
    \item Extensive experiments spanning domains from \textbf{general mathematical reasoning} to \textbf{specialized medical diagnostics} demonstrate that DVRP significantly outperforms state-of-the-art RLVR methods like GRPO~\cite{shao2024deepseekmathpushinglimitsmathematical} and DAPO~\cite{dapo}, effectively enhancing visual perception capabilities.
\end{itemize}

\section{Related Work}

\subsection{Multimodal Reasoning}
Recent advancements in Multimodal Large Language Models (MLLMs) have significantly extended the reasoning capabilities of LLMs to visual domains \cite{wang2024comprehensive}. Pioneering works such as LLaVA~\cite{liu2024visual} and Qwen-VL~\cite{bai2023qwen} demonstrated that visual instruction tuning could effectively align visual encoders with LLMs, enabling strong performance on general visual-oriented benchmarks~\cite{yue2024mmmu}. Following this, the community has focused on enhancing the reasoning depth of these models. Inspired by the efficacy of CoT in LLMs, the integration of Multimodal-CoT~\cite{zhang2024multimodalchainofthoughtreasoninglanguage, sarch2025grounded} and the curation of high-quality reasoning datasets~\cite{chen2024sharegpt4v, chen2023shikra, dong2025insight, sunchiron} have emerged as pivotal strategies for bolstering visual understanding, thereby significantly mitigating linguistic hallucinations~\cite{huang2024opera, leng2024mitigating}.

To substantially enhance visual understanding, a novel paradigm, which termed \textit{visual reasoning in action} has emerged, advocating for thinking with images by explicitly incorporating visual content within CoT processes~\cite{su2025thinking}. Representative strategies employ visual programming via code generation~\cite{suris2023vipergpt, gupta2023visual, lin2025vcodemultimodalcodingbenchmark}, leverage external visual utilities (e.g., crop, zoom, rotate)~\cite{zheng2025deepeyes, zhang2025thinking,qi2024cogcom, liu2023llavapluslearningusetools}, or orchestrate diverse expert models to tackle multimodal tasks~\cite{shen2023hugginggpt,lu2023chameleon}. 
Further explorations extend to tool-integrated, multi-turn, and multi-agent system for interleaved reasoning~\cite{wu2025vtool, chen2024agentverse, wang2025vgr, man2025argus} and unified frameworks~\cite{han2025controlthinker, li2025zebra}. 
\textit{However, these methods rely heavily on external sources such as expensive trajectory data, agentic tools, or advanced teacher models. This incurs significant overhead and does not fundamentally improve the model's intrinsic visual perception and reasoning capabilities.}

\subsection{Advancements in RLVR Frameworks}
Recent research on RLVR focuses on optimizing three critical components: \textbf{Data}, \textbf{Reward}, and \textbf{Rollout}.

\noindent\textbf{Data Curation.} 
High-quality Chain-of-Thought initialization is a prerequisite for stable RL training~\cite{chen2025acereason, vision-r1, r1-vl, modomodo}. Complementing this, recent studies employ dynamic data selection strategies, such as filtering trajectories based on value estimation~\cite{thinklite} or applying perplexity- and difficulty-based correction mechanisms~\cite{kong2024perplexityaware, capo} to enhance sample efficiency and diversity.

\noindent\textbf{Reward Engineering.} 
Granular reward design serves as a pivotal mechanism in RLVR~\cite{deepseekr1}. Dominant approaches rely on \textit{outcome-based} signals, including rigorous accuracy verification~\cite{liu2025visualagenticreinforcementfinetuning}, strict formatting and length constraints~\cite{parthasarathi2025grpolambdacreditassignmentimproves,zhang2025grpoleaddifficultyawarereinforcementlearning}, caption alignment metrics~\cite{RACRO-7B-CRO} for visual-semantic understanding, and grounding constraints~\cite{zhang2025medtvtr1multimodalllmempowering}. To further mitigate reasoning shortcuts, recent methods augment these with \textit{process-level constraints}~\cite{capo}, aiming to bolster the procedural correctness and reliability of intermediate reasoning chains.

\noindent\textbf{Rollout Optimization.} 
Enhancing the efficiency of the exploration phase is another active frontier. Recent strategies focus on optimizing the rollout space by dynamically adjusting sampling temperatures~\cite{liao2025enhancingefficiencyexplorationreinforcement}, expanding the sampling space~\cite{noisyrollout,li2025revisitingvisualunderstandingmultimodal}, polishing reasoning processes~\cite{fan2025sophiavlr1reinforcingmllmsreasoning}, and employing tree-search algorithms to diversify generation paths~\cite{ji2025treesearch}. These methods aim to balance exploration and exploitation, ensuring the policy covers a broader solution space without diverging into incoherence.

\textit{\noindent However, these strategies relegate the visual modality to an auxiliary role relative to text, neglecting the critical need to enhance visual robustness and sensitivity in MLLMs. 
In contrast, our proposed DVRP intrinsically fosters the model's visual perception capabilities.}


\begin{figure*}[h]
  \includegraphics[width=\linewidth]{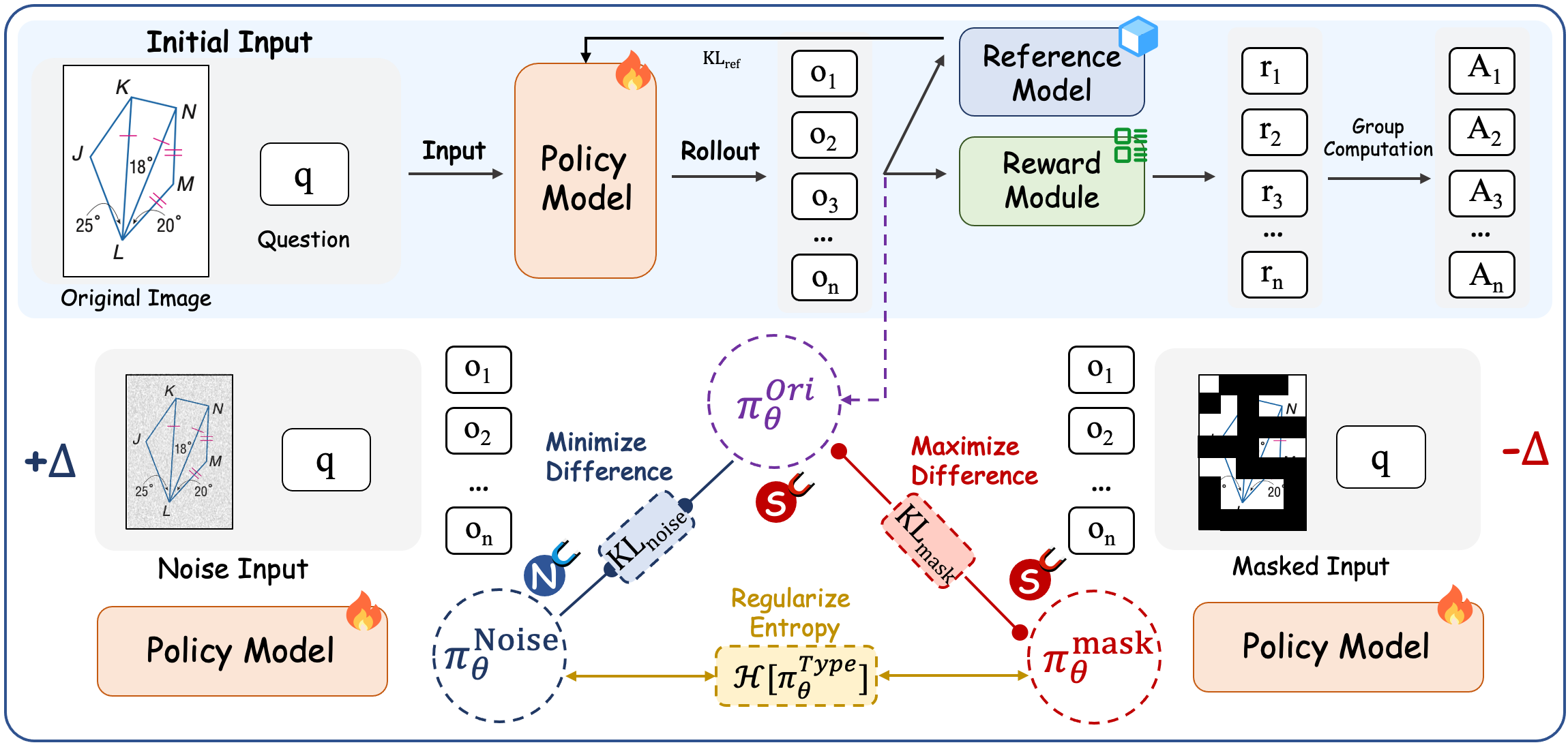}
  \caption{The framework of DVRP. Our method bridges the perception-reasoning decoupling via a visual triplet contrastive learning objective. The upper stream represents the standard reasoning rollout. The lower streams enforce two critical visual properties: (1) Visual Robustness: minimizing the KL-divergence ($KL_{noise}$) between predictions on original and noise-perturbed inputs ($+\Delta$); and (2) Visual Sensitivity: maximizing the divergence ($KL_{mask}$) when critical visual semantics are occluded ($-\Delta$). An entropy regularization term $\mathcal{H}$ prevents distribution collapse.}
  \label{fig:pipeline}
\end{figure*}

\section{Method}

\subsection{Problem Formulation}

\paragraph{Preliminaries on GRPO.}
We consider a multimodal reasoning task where a policy model $\pi_\theta$, parameterized by $\theta$, takes a visual input $I$ and a textual query $q$ to generate a reasoning chain followed by a final answer, denoted as $o$. The training dataset is represented as $\mathcal{D} = \{(I, q, a)\}$, where $a$ is the ground truth answer.

Recent advancements in RLVR have largely adopted GRPO~\cite{shao2024deepseekmathpushinglimitsmathematical} to enhance reasoning capabilities without the need for a separate value network. 
Formally, for each input instance $(I, q)$, GRPO samples a group of $G$ outputs $\{o_i\}_{i=1}^G$ from the old policy $\pi_{\theta_{old}}$. A rule-based reward function $r(o, a)$ evaluates the correctness of each output (e.g., format compliance and answer accuracy), assigning a reward $R_i$. To reduce variance, the advantage $\hat{A}_i$ for the $i$-th output is computed by normalizing the rewards within the group:
\begin{equation}
    \hat{A}_i = \frac{R_i - \text{mean}(\{R_j\}_{j=1}^G)}{\text{std}(\{R_j\}_{j=1}^G) + \epsilon},
\end{equation}
where $\epsilon$ is a small constant for numerical stability. The optimization objective of GRPO is defined as:
\begin{equation}
    \mathcal{J}_{\text{GRPO}}(\theta) = \mathbb{E}_{q \sim \mathcal{D}, \{o_i\} \sim \pi_{\theta_{old}}} \left[ \frac{1}{G} \sum_{i=1}^G \mathcal{L}_{\text{clip}}(o_i, \hat{A}_i) \right],
\end{equation}
where $\mathcal{L}_{\text{clip}}$ denotes the standard PPO clipped surrogate loss:
\begin{equation}
    \mathcal{L}_{\text{clip}} = \min \left( \rho_i \hat{A}_i, \text{clip}(\rho_i, 1-\epsilon_{clip}, 1+\epsilon_{clip}) \hat{A}_i \right),
\end{equation}
with $\rho_i = \frac{\pi_\theta(o_i | I, q)}{\pi_{\theta_{old}}(o_i | I, q)}$ representing the probability ratio. Following recent practices~\cite{dapo}, we remove the KL divergence penalty term to encourage sufficient exploration.

\paragraph{The Perception-Reasoning Decoupling.}
A critical limitation of current multimodal RLVR paradigms is their failure to enforce the \textit{causal dependency} between visual perception and reasoning paths~\cite{capo}. By relying solely on outcome-based textual rewards and reasoning in language medium, these methods treat visual inputs as optional context rather than necessary evidence. Consequently, existing frameworks largely overlook the model's \textbf{sensitivity to critical visual semantics} and \textbf{robustness against visual perturbations}, inadvertently incentivizing linguistic shortcuts over visually grounded reasoning~\cite{han2025learningseeingdemystifyingllm}.

\subsection{Visual Triplets as Intrinsic Supervision}
To bridge this gap, we propose \textbf{DVRP}, which introduces intrinsic supervision via a \textit{Visual Triplet} mechanism. As illustrated in Figure~\ref{fig:pipeline} , for each training instance, we construct three distinct views:

\begin{itemize}
    \item \textbf{Invariant View ($I$):} The original visual input, serving as the anchor for standard reasoning. The policy distribution under this view is denoted as $\pi_\theta^{Ori} = \pi_\theta(\cdot | I, q)$.
    \item \textbf{Decremental View ($I_{mask}$):} A ``$-\Delta$'' state generated by masking visual regions randomly. This creates a counterfactual scenario where visual evidence is lost. The corresponding policy is $\pi_\theta^{Mask} = \pi_\theta(\cdot | I_{mask}, q)$.
    \item \textbf{Incremental View ($I_{noise}$):} A ``$+\Delta$'' state generated by injecting non-semantic Diffusion noise. This simulates environmental instability while preserving semantics. The corresponding policy is $\pi_\theta^{Noise} = \pi_\theta(\cdot | I_{noise}, q)$.
\end{itemize}
This triplet structure allows us to operationalize two complementary learning signals: \textit{Visual Sensitivity} (diverging from $\pi_\theta^{Mask}$) and \textit{Visual Robustness} (aligning with $\pi_\theta^{Noise}$). We conducted comprehensive experiments to investigate the insights and applicability of our approach, which are detailed in Section~\ref{sec:robustness_analysis}.

\begin{table*}[t]
\centering
\resizebox{\textwidth}{!}{%
\begin{tabular}{l | cccccc cc | ccccc c | cc}
\toprule
\multirow{2}{*}{\textbf{Method}} & 
\multicolumn{8}{c|}{\cellcolor{generalbg}\textbf{General Multimodal Reasoning}} & 
\multicolumn{6}{c|}{\cellcolor{medicalbg}\textbf{Medical Multimodal Reasoning}} & 
\multicolumn{2}{c}{\cellcolor{overallbg}\textbf{Overall}} \\
\cmidrule{2-17}
 & \textbf{Geo3k} & \textbf{Vista} & \textbf{WeMath} & \textbf{MVerse} & \textbf{MVerse-V} & \textbf{MMKI2} & \textbf{AVG} & \textbf{$\Delta_{rel}$} & \textbf{Slake} & \textbf{Path} & \textbf{Rad} & \textbf{PMC} & \textbf{AVG} & \textbf{$\Delta_{rel}$} & \textbf{AVG} & \textbf{$\Delta_{rel}$} \\
\midrule
\multicolumn{17}{c}{\textit{Qwen2.5-VL-3B Backbone}} \\
\midrule
\rowcolor{gray!10}
\textbf{Base Model} & 20.6 & 40.6 & 23.9 & 30.9 & 28.2 & 34.8 & 29.8 & \textit{Ref.} & 48.7 & 59.2 & 40.3 & 46.8 & 48.7 & \textit{Ref.} & 37.4 & \textit{Ref.} \\
GRPO          & 28.7 & 59.3 & 58.9 & 55.3 & 52.2 & 57.2 & 51.9 & \cellcolor{gainbg}\textcolor{gaintext}{$\uparrow$74.2\%} & 70.9 & 74.5 & 71.2 & 53.4 & 67.5 & \cellcolor{gainbg}\textcolor{gaintext}{$\uparrow$38.6\%} & 58.2 & \cellcolor{gainbg}\textcolor{gaintext}{$\uparrow$55.6\%} \\
DAPO          & 31.2 & 60.9 & 60.0 & 56.3 & 53.0 & \textbf{66.8} & 54.7 & \cellcolor{gainbg}\textcolor{gaintext}{$\uparrow$83.6\%} & 71.3 & 72.4 & 70.8 & 56.6 & 67.8 & \cellcolor{gainbg}\textcolor{gaintext}{$\uparrow$39.2\%} & 59.9 & \cellcolor{gainbg}\textcolor{gaintext}{$\uparrow$60.2\%} \\
NoiseRollout  & 32.5 & 63.0 & 60.1 & 56.9 & 53.5 & 63.4 & 54.9 & \cellcolor{gainbg}\textcolor{gaintext}{$\uparrow$84.2\%} & 72.5 & 76.1 & 75.0 & 61.0 & 71.2 & \cellcolor{gainbg}\textcolor{gaintext}{$\uparrow$46.2\%} & 61.4 & \cellcolor{gainbg}\textcolor{gaintext}{$\uparrow$64.2\%} \\
PAPO          & 30.9 & 61.3 & 60.1 & 57.1 & 53.9 & 57.3 & 53.4 & \cellcolor{gainbg}\textcolor{gaintext}{$\uparrow$79.2\%} & 72.1 & 75.9 & 74.8 & 60.7 & 70.9 & \cellcolor{gainbg}\textcolor{gaintext}{$\uparrow$45.6\%} & 60.4 & \cellcolor{gainbg}\textcolor{gaintext}{$\uparrow$61.5\%} \\
\noalign{\vskip 2pt}
\cdashline{1-17}
\noalign{\vskip 3pt}
\textbf{DVRP$_D$ (Ours)}   & \textbf{35.1} & 64.9 & \textbf{60.5} & \textbf{58.1} & \textbf{54.8} & 60.9 & \textbf{55.7} & \cellcolor{gainbg}\textcolor{gaintext}{\textbf{$\uparrow$86.9\%}} & 76.3 & \textbf{78.9} & 75.9 & \textbf{62.2} & \textbf{73.3} & \cellcolor{gainbg}\textcolor{gaintext}{\textbf{$\uparrow$50.5\%}} & \textbf{62.7} & \cellcolor{gainbg}\textcolor{gaintext}{\textbf{$\uparrow$67.6\%}} \\
\textbf{DVRP$_G$ (Ours)} & 34.5 & \textbf{65.5} & 60.3 & 57.7 & 54.5 & 61.2 & 55.6 & \cellcolor{gainbg}\textcolor{gaintext}{$\uparrow$86.6\%} & \textbf{74.1} & 77.5 & \textbf{76.2} & 61.5 & 72.3 & \cellcolor{gainbg}\textcolor{gaintext}{$\uparrow$48.5\%} & 62.3 & \cellcolor{gainbg}\textcolor{gaintext}{$\uparrow$66.6\%} \\
\midrule
\multicolumn{17}{c}{\textit{Qwen2.5-VL-7B Backbone}} \\
\midrule
\rowcolor{gray!10}
\textbf{Base Model} & 33.8 & 55.9 & 41.8 & 45.6 & 36.9 & 43.7 & 43.0 & \textit{Ref.} & 63.7 & 60.6 & 61.3 & 52.2 & 59.5 & \textit{Ref.} & 49.6 & \textit{Ref.} \\
GRPO          & 40.2 & 65.5 & 66.1 & 66.5 & 61.7 & 72.1 & 62.0 & \cellcolor{gainbg}\textcolor{gaintext}{$\uparrow$44.2\%} & 74.7 & 75.7 & 74.9 & 56.2 & 70.4 & \cellcolor{gainbg}\textcolor{gaintext}{$\uparrow$18.3\%} & 65.4 & \cellcolor{gainbg}\textcolor{gaintext}{$\uparrow$31.9\%} \\
DAPO          & 35.9 & 61.9 & 58.5 & 55.6 & 51.0 & 71.9 & 55.8 & \cellcolor{gainbg}\textcolor{gaintext}{$\uparrow$29.8\%} & 74.9 & 76.2 & 67.8 & 57.2 & 69.0 & \cellcolor{gainbg}\textcolor{gaintext}{$\uparrow$16.0\%} & 61.1 & \cellcolor{gainbg}\textcolor{gaintext}{$\uparrow$23.2\%} \\
NoiseRollout  & 39.7 & 67.8 & 65.3 & 66.1 & 62.8 & 70.5 & 62.0 & \cellcolor{gainbg}\textcolor{gaintext}{$\uparrow$44.2\%} & 72.8 & 74.1 & 72.2 & 59.9 & 69.8 & \cellcolor{gainbg}\textcolor{gaintext}{$\uparrow$17.3\%} & 65.1 & \cellcolor{gainbg}\textcolor{gaintext}{$\uparrow$31.3\%} \\
PAPO          & 40.2 & 69.5 & 66.7 & 68.4 & 64.9 & 72.5 & 63.7 & \cellcolor{gainbg}\textcolor{gaintext}{$\uparrow$48.1\%} & 76.9 & 73.4 & 77.5 & 61.6 & 72.4 & \cellcolor{gainbg}\textcolor{gaintext}{$\uparrow$21.7\%} & 67.2 & \cellcolor{gainbg}\textcolor{gaintext}{$\uparrow$35.5\%} \\
\noalign{\vskip 2pt}
\cdashline{1-17}
\noalign{\vskip 3pt}
\textbf{DVRP$_D$ (Ours)}          & \textbf{43.4} & 70.9 & 67.8 & \textbf{68.9} & 65.3 & 74.1 & 65.1 & \cellcolor{gainbg}\textcolor{gaintext}{$\uparrow$51.4\%} & 80.3 & \textbf{79.7} & \textbf{80.5} & 64.2 & 76.2 & \cellcolor{gainbg}\textcolor{gaintext}{$\uparrow$28.1\%} & 69.5 & \cellcolor{gainbg}\textcolor{gaintext}{$\uparrow$40.1\%} \\
\textbf{DVRP$_G$ (Ours)} & 42.3 & \textbf{71.1} & \textbf{68.1} & 67.4 & \textbf{66.7} & \textbf{75.6} & \textbf{65.2} & \cellcolor{gainbg}\textcolor{gaintext}{\textbf{$\uparrow$51.6\%}} & \textbf{81.7} & 78.3 & 79.9 & \textbf{65.5} & \textbf{76.4} & \cellcolor{gainbg}\textcolor{gaintext}{\textbf{$\uparrow$28.4\%}} & \textbf{69.7} & \cellcolor{gainbg}\textcolor{gaintext}{\textbf{$\uparrow$40.5\%}} \\
\bottomrule
\end{tabular}%
}
\caption{Performance (\textbf{avg@8} acc \%) comparison of Qwen2.5-VL-3B and 7B backbones. We compare our \textbf{DVRP} against baselines including GRPO, DAPO, and PAPO. Subscripts $_D$ and $_G$ denote optimizations using DAPO and GRPO, respectively. The \textbf{Base Model} (highlighted) serves as the reference for calculating relative improvements ($\Delta_{rel} = \frac{\text{Method}-\text{Base}}{\text{Base}} \times 100\%$). For a fair comparison, all baselines are reproduced in the same environment and evaluated using the identical suite.}
\label{tab:main_results}
\end{table*}

\subsection{Optimization Objective}
Formally, DVRP optimizes a unified objective that balances task performance with intrinsic visual grounding. We integrate visual triplet supervision into the GRPO framework \textbf{to enforce} two complementary constraints:
(1) \textbf{Visual Sensitivity:} The policy must diverge from the decremental view ($\pi_{\theta}^{Mask}$) where visual evidence is absent;
(2) \textbf{Visual Robustness:} The policy must remain consistent with the incremental view ($\pi_{\theta}^{Noise}$) despite perturbations.

However, directly constraining the consistency in the robustness and sensitivity branches can lead to a degenerate solution where the policy collapses into a high-entropy uniform distribution to trivially minimize the KL divergence. To mitigate this risk, we incorporate an entropy regularization term to penalize high uncertainty, preventing the model from converging to such a trivial state. Consequently, the final optimization objective $\mathcal{J}_{\text{DVRP}}$ is formulated as:
\begin{equation}
\begin{aligned}
    \mathcal{J}_{\text{DVRP}}(\theta) = & \underbrace{\mathcal{J}_{\text{GRPO}}(\theta)}_{\text{Maximize Reward}} \\
    & \underbrace{+ \lambda_{nec} \cdot \mathbb{D}_{KL}\left( \pi_\theta^{Ori} \parallel \pi_{\theta}^{Mask} \right)}_{\text{\textcolor{red}{Visual Sensitivity} (Max Difference)}} \\
    & \underbrace{- \lambda_{rob} \cdot \mathbb{D}_{KL}\left( \pi_\theta^{Ori} \parallel \pi_{\theta}^{Noise} \right)}_{\text{\textcolor{blue}{Visual Robustness} (Min Difference)}} \\
    & \underbrace{- \lambda_{ent} \cdot \mathbb{E}\left[\mathcal{H}(\pi_\theta^{Noise}) + \mathcal{H}(\pi_\theta^{Mask})\right]}_{\text{\textcolor{orange}{Entropy Penalty} (Prevent Collapse)}},
\end{aligned}
\end{equation}
where $\pi_{\theta_{old}}$ (implicitly used in $\mathcal{J}_{\text{GRPO}}$) serves as the frozen reference distribution to stabilize the reward maximization, while the triplet objectives enforce constraints on the current policy manifold.

\begin{table*}[t!]
\centering
\small
\setlength{\tabcolsep}{4.5pt} 
\renewcommand{\arraystretch}{1.15} 

\definecolor{deltagreen}{HTML}{008000} 
\newcommand{\inc}[1]{\fontsize{7pt}{7pt}\selectfont\textcolor{deltagreen}{\textbf{+#1}}}

\resizebox{\textwidth}{!}{%
\begin{tabular}{l | cccc | cc}
\toprule

\multirow{2.5}{*}{\textbf{Method}} & 
\multicolumn{4}{c|}{\cellcolor{generalbg}\textbf{In-Domain Datasets}} & 
\multicolumn{2}{c}{\cellcolor{medicalbg}\textbf{Out-of-Domain Datasets}} \\
\cmidrule{2-7}
 & \textbf{PMCVQA} & \textbf{VQA-RAD} & \textbf{SLAKE} & \textbf{PathVQA} & \textbf{MedXpertQA} & \textbf{MMMU-Med} \\
\midrule

\multicolumn{7}{c}{\textit{Proprietary Models}} \\
\midrule
Gemini-2.0-flash-lite & 50.8 & 59.4 & 73.1 & 64.9 & -- & 58.7 \\
GPT-4.1-Nano          & 53.1 & 61.8 & 73.1 & 70.6 & -- & 60.6 \\
GPT-4o                & --   & 63.9 & 71.6 & 75.9 & -- & -- \\

\midrule
\multicolumn{7}{c}{\textit{General-purpose Multimodal VLMs}} \\
\midrule
Qwen-VL-Chat~\cite{qwen2-vl}        & 36.6 & 47.0 & 56.0 & 55.1 & -- & 32.7 \\
Yi-VL-34B~\cite{yi-vl}              & 39.5 & 53.0 & 58.9 & 47.3 & -- & 41.5 \\
LLaVA-v1.6-7B~\cite{llava-v1.6}     & 35.5 & 52.6 & 57.9 & 47.9 & -- & 33.1 \\
LLaVA-v1.6-13B~\cite{llava-v1.6}    & 36.6 & 55.8 & 58.9 & 51.9 & -- & 39.3 \\
LLaVA-v1.6-34B~\cite{llava-v1.6}    & 44.4 & 58.6 & 67.3 & 59.1 & -- & 48.8 \\
LLaVA-v1.5-LLaMA3-8B~\cite{llava-v1.5-llama} & 36.4 & 54.2 & 59.4 & 54.1 & -- & 38.2 \\

\midrule
\multicolumn{7}{c}{\textit{Medical Multimodal VLMs}} \\
\midrule
Med-Flamingo~\cite{medflamingo}     & 23.3 & 45.4 & 43.5 & 54.7 & 22.1 & 28.3 \\
RadFM~\cite{radfm}                  & 25.9 & 50.6 & 34.6 & 38.7 & 23.4 & 27.0 \\
LLaVA-Med-7B~\cite{llavamed}        & 24.7 & 51.4 & 48.6 & 56.8 & 20.8 & 36.9 \\
LLaVA\_Med-LLaMA3-8B~\cite{llava-v1.5-llama} & 46.6 & 60.2 & 61.2 & 54.5 & -- & 41.1 \\
PubMedVision-8B~\cite{huatuo}       & 52.7 & 63.8 & 74.5 & 59.9 & -- & 49.1 \\
HuatuoGPT-Vision-34B~\cite{huatuo}  & 58.2 & 68.1 & 76.9 & 63.5 & 22.1 & 54.4 \\
MedVLThinker-7B~\cite{medvlthinker} & 57.5 & 63.7 & 67.8 & 65.2 & 20.9 & 57.0 \\
CAPO-7B~\cite{capo}                 & 55.5 & 78.5 & 79.1 & 68.9 & -- & 60.0 \\

\midrule
\multicolumn{7}{c}{\textit{Medical Agentic Systems}} \\
\midrule
MedAgents~\cite{medagents}          & -- & 65.6 & 67.9 & 63.2 & -- & 49.7 \\
MDAgents~\cite{mdagents}            & -- & 66.8 & 68.2 & 65.4 & -- & 52.3 \\
AFlow~\cite{aflow}                  & -- & 67.3 & 68.9 & 66.4 & -- & 53.6 \\
MMedAgent-RL-7B~\cite{mmedagents}   & -- & 71.5 & 76.2 & 72.3 & -- & \textbf{66.4} \\

\midrule
\multicolumn{7}{c}{\textit{\textbf{Ours}}} \\
\midrule

\rowcolor{graybg}
\textbf{Base Model} \textit{(Qwen2.5-VL-3B)} & 46.8 & 40.3 & 48.7 & 59.2 & 20.7 & 31.5 \\

\rowcolor{gainbg} 
~+ DVRP$_G$           & 61.5 & 76.2 & 74.1 & 77.5 & 23.2 & 42.4 \\[-0.2em]
\rowcolor{gainbg}
                      & \inc{14.7} & \inc{35.9} & \inc{25.4} & \inc{18.3} & \inc{2.5} & \inc{10.9} \\

\rowcolor{gainbg} 
~+ DVRP$_D$           & 62.2 & 75.9 & 76.3 & \underline{78.9} & \underline{24.1} & 44.1 \\[-0.2em]
\rowcolor{gainbg}
                      & \inc{15.4} & \inc{35.6} & \inc{27.6} & \inc{19.7} & \inc{3.4} & \inc{12.6} \\

\noalign{\vskip 2pt}
\cdashline{1-7}
\noalign{\vskip 3pt}

\rowcolor{graybg}
\textbf{Base Model} \textit{(Qwen2.5-VL-7B)} & 52.2 & 61.3 & 63.7 & 60.6 & 20.1 & 54.7 \\

\rowcolor{gainbg} 
~+ DVRP$_G$           & \textbf{65.5} & \underline{79.9} & \textbf{81.7} & 78.3 & 24.0 & \underline{66.2} \\[-0.2em]
\rowcolor{gainbg}
                      & \inc{13.3} & \inc{18.6} & \inc{18.0} & \inc{17.7} & \inc{3.9} & \inc{11.5} \\

\rowcolor{gainbg} 
~+ DVRP$_D$           & \underline{64.2} & \textbf{80.5} & \underline{80.3} & \textbf{79.7} & \textbf{25.7} & 65.9 \\[-0.2em]
\rowcolor{gainbg}
                      & \inc{12.0} & \inc{19.2} & \inc{16.6} & \inc{19.1} & \inc{5.6} & \inc{11.2} \\

\bottomrule
\end{tabular}%
}
\caption{Comprehensive performance comparison on 2D medical VQA benchmarks. \textbf{Bold} and \underline{underline} denote best and second-best scores. The green values indicate the absolute improvement over the corresponding Base Model. Subscripts $_D$ and $_G$ denote optimizations using DAPO and GRPO, respectively. }
\label{tab:medical_results}
\end{table*}

\begin{table*}[t!]
\centering
\small
\setlength{\tabcolsep}{4.5pt}
\renewcommand{\arraystretch}{1.15}

\definecolor{deltagreen}{HTML}{008000} 
\newcommand{\inc}[1]{\fontsize{7pt}{7pt}\selectfont\textcolor{deltagreen}{\textbf{+#1}}}

\resizebox{\textwidth}{!}{%
\begin{tabular}{l ccccccc}
\toprule

\multirow{2.5}{*}{\textbf{Method}} & 
\multicolumn{7}{c}{\cellcolor{generalbg}\textbf{Mathematical Multimodal Reasoning}} \\
\cmidrule{2-8} 
 & \textbf{Geo3k} & \textbf{Vista} & \textbf{WeMath} & \textbf{MVerse} & \textbf{MVerse-V} & \textbf{MMKI2} & \textbf{AVG} \\
\midrule

\multicolumn{8}{c}{\textit{Proprietary Models}} \\
\midrule
GPT-4o                & -- & 64.7 & 62.8 & 50.2 & 53.8 & 55.8 & 57.5 \\
GPT-4o-mini           & -- & 59.9 & 56.3 & 42.3 & 45.1 & 51.9 & 51.1 \\
Gemini-2.0-flash      & -- & 70.4 & 47.4 & 47.8 & 48.7 & 65.2 & 55.9 \\

\midrule
\multicolumn{8}{c}{\textit{General-purpose Multimodal VLMs}} \\
\midrule
Qwen2.5-VL-72B~\cite{qwen2-vl}        & -- & \textbf{74.2} & 49.1 & 47.3 & 48.6 & 70.5 & 57.9 \\
InternVL2.5-8B~\cite{internvl2.5}     & -- & 64.9 & 44.9 & 37.0 & 40.2 & 46.8 & 46.8 \\
InternVL2.5-VL-78B ~\cite{internvl2.5} & -- & 64.9 & 44.9 & 37.0 & 40.2 & 59.8 & 49.4 \\
LLaVA-OneVision-7B~\cite{llava-onevision} & -- & 58.5 & 44.1 & -- & -- & -- & 51.3 \\
LLaVA-OneVision-72B~\cite{llava-onevision}  & -- & 67.1 & 32.0 & 27.2 & 30.1 & -- & 39.1 \\
LLaVA-OneVision-1.5-8B~\cite{llava-onevision-1.5} & -- & 69.6 & 61.5 & -- & -- & -- & \textbf{65.6} \\
LLaVA-Critic-R1-7B~\cite{llava-critic}   & 35.4 & 68.7 & 62.6 & 58.9 & 53.1 & 57.4 & 56.0 \\
R1-OneVision-7B~\cite{r1onevision}     & 30.6 & 64.9 & 55.2 & 61.7 & 44.3 & 43.3 & 50.0 \\

\midrule
\multicolumn{8}{c}{\textit{Math-Specific Multimodal VLMs}} \\
\midrule
MM-Eureka-7B~\cite{mm-eureka}               & 36.4 & 59.1 & 45.3 & 57.6 & 56.4 & 60.6 & 52.6 \\
MM-Eureka-7B-CPGD~\cite{mm-euruka-cpgd}     & 37.6 & 64.2 & 64.3 & 63.7 & 59.2 & 64.7 & 59.0 \\
ADORA-7B~\cite{gui2025adora}                & 41.2 & 61.1 & 53.0 & 45.2 & 41.8 & 49.8 & 48.7 \\
R1-VL-7B~\cite{r1-vl}                       & 31.9 & 63.5 & 56.1 & 42.0 & 43.2 & 55.3 & 48.7 \\
VLAA-Thinker-7B~\cite{vlaa-thinker}         & 24.2 & 67.4 & 65.9 & 47.9 & 52.0 & 63.2 & 53.4 \\
VL-Rethinker-7B~\cite{vl-rethinker}         & 33.6 & 61.3 & 66.5 & 64.0 & 60.8 & 59.8 & 57.7 \\
RACRO-7B-CRO-GRPO~\cite{RACRO-7B-CRO}       & 41.4 & 61.7 & \textbf{68.9} & 65.7 & 61.7 & 70.5 & 61.7 \\
ThinkLite-7B-VL~\cite{thinklite}            & 34.4 & 68.9 & 63.5 & 49.5 & 46.0 & 56.2 & 53.1 \\

\midrule
\multicolumn{8}{c}{\textit{\textbf{Ours}}} \\
\midrule

\rowcolor{graybg}
\textbf{Base Model} \textit{(Qwen2.5-VL-3B)} & 20.6 & 40.6 & 23.9 & 30.9 & 28.2 & 34.8 & 29.8 \\

\rowcolor{gainbg}
~+ DVRP$_D$                 & 35.1 & 64.9 & 60.5 & 58.1 & 54.8 & 60.9 & 55.7 \\[-0.2em]
\rowcolor{gainbg}
                            & \inc{14.5} & \inc{24.3} & \inc{36.6} & \inc{27.2} & \inc{26.6} & \inc{26.1} & \inc{25.9} \\

\rowcolor{gainbg}
~+ DVRP$_G$                 & 34.5 & 65.5 & 60.3 & 57.7 & 54.5 & 61.2 & 55.6 \\[-0.2em]
\rowcolor{gainbg}
                            & \inc{13.9} & \inc{24.9} & \inc{36.4} & \inc{26.8} & \inc{26.3} & \inc{26.4} & \inc{25.8} \\

\noalign{\vskip 2pt}
\cdashline{1-8}
\noalign{\vskip 3pt}

\rowcolor{graybg}
\textbf{Base Model} \textit{(Qwen2.5-VL-7B)} & 33.8 & 55.9 & 41.8 & 45.6 & 36.9 & 43.7 & 43.0 \\

\rowcolor{gainbg}
~+ DVRP$_D$                 & \textbf{43.4} & 70.9 & 67.8 & \textbf{68.9} & \underline{65.3} & \underline{74.1} & 65.1 \\[-0.2em]
\rowcolor{gainbg}
                            & \inc{9.6} & \inc{15.0} & \inc{26.0} & \inc{23.3} & \inc{28.4} & \inc{30.4} & \inc{22.1} \\

\rowcolor{gainbg}
~+ DVRP$_G$                 & \underline{42.3} & \underline{71.1} & \underline{68.1} & \underline{67.4} & \textbf{66.7} & \textbf{75.6} & \underline{65.2} \\[-0.2em]
\rowcolor{gainbg}
                            & \inc{8.5} & \inc{15.2} & \inc{26.3} & \inc{21.8} & \inc{29.8} & \inc{31.9} & \inc{22.2} \\

\bottomrule
\end{tabular}%
}
\caption{Performance comparison on mathematical multimodal reasoning benchmarks. \textbf{Bold} and \underline{underline} denote the best and second-best performance, respectively. The green values indicate the absolute improvement over the corresponding Base Model. Subscripts $_D$ and $_G$ denote optimizations using DAPO and GRPO, respectively.}
\label{tab:math_results}
\end{table*}

\section{Experiments}
\subsection{Experiment Setup and Training Details}
\noindent \textbf{Implementation Framework.} Our experimental framework is built upon Easy-R1~\cite{zheng2025easyr1}, implemented using Python 3.10 and PyTorch 2.4.0~\cite{torch} with CUDA 12.4 support. Following the training paradigm established by DeepSeek-R1~\cite{deepseekr1}, we employed Direct Reinforcement Fine-Tuning (RFT) to optimize the Qwen2.5-VL-7B and 3B backbones~\cite{qwen2.5}. All experiments were conducted on a computational cluster equipped with $4 \times$ NVIDIA A800 GPUs.

\vspace{0.5em}
\noindent \textbf{Training Datasets.} To foster robust reasoning across domains, we employ ViRL39K~\cite{vl-rethinker} for mathematical training, while for the medical domain, we construct a composite dataset by amalgamating the training splits of Slake~\cite{slake}, PathVQA~\cite{pathvqa}, RadVQA~\cite{radvqa}, and PMC-VQA~\cite{pmcvqa}.

\vspace{0.5em}
\noindent \textbf{Training Details.} To demonstrate the versatility of the proposed \textit{DVRP}, we integrated DVRP with two fundamental RLVR algorithms, DAPO~\cite{dapo} and GRPO~\cite{shao2024deepseekmathpushinglimitsmathematical}, and evaluated its impact on the Qwen2.5-VL-7B and 3B backbones~\cite{qwen2.5}. We conducted extensive ablation studies to determine the optimal masking ratios and noise levels~\cite{capo}. Details are provided in Appendix~\ref{sec:pertubation_ablation}. Following the methodology of NoiseRollout~\cite{noisyrollout}, we adopted a sigmoid function for diffusion noise annealing (Detailed in Section~\ref{sec:noise}).
We also adopt random patch masking as the visual sensitivity operation in this work, a choice that has been discussed in~\cite{papo}.
Specifically, the KL divergence terms are computed as the summation of \textbf{token-level KL divergences} between the categorical output distributions of the compared policies at each generation step, calculated over the trajectories sampled from the invariant policy $\pi_\theta^{Ori}$.
Detailed hyperparameter configurations are provided in Appendix~\ref{sec:hyperparameter}.

\vspace{0.5em}
\noindent \textbf{Evaluation Protocol.} We conduct a comprehensive evaluation across benchmarks spanning both in-domain and out-of-domain settings. Our comparative analysis includes a wide range of baselines, comprising closed-source commercial models, open-source general-purpose models, and domain-specific reasoning models. For detailed experimental settings, please refer to Section~\ref{sec:evaluation}

\begin{table*}[t!]
\centering
\resizebox{\textwidth}{!}{%
\begin{tabular}{l | cccccc c | cccc c | c}
\toprule
\multirow{2}{*}{\textbf{Method}} & 
\multicolumn{7}{c|}{\cellcolor{generalbg}\textbf{General Multimodal Reasoning}} & 
\multicolumn{5}{c|}{\cellcolor{medicalbg}\textbf{Medical Multimodal Reasoning}} & 
\cellcolor{overallbg}\textbf{Overall} \\
\cmidrule{2-14}
 & \textbf{Geo3k} & \textbf{Vista} & \textbf{WeMath} & \textbf{MVerse} & \textbf{MVerse-V} & \textbf{MMKI2} & \textbf{AVG} & \textbf{Slake} & \textbf{Path} & \textbf{Rad} & \textbf{PMC} & \textbf{AVG} & \textbf{AVG} \\
\midrule
\multicolumn{14}{c}{\textit{Qwen2.5-VL-7B Backbone}} \\
\midrule
\rowcolor{gray!10}
Base Model (Original) & 33.8 & 55.9 & 41.8 & 45.6 & 36.9 & 43.7 & 43.0 & 63.7 & 60.6 & 61.3 & 52.2 & 59.5 & 49.6 \\
\noalign{\vskip 2pt}
Baseline (GRPO) & 40.2 & 65.5 & 66.1 & 66.5 & 61.7 & 72.1 & 62.0 & 74.7 & 75.7 & 74.9 & 56.2 & 70.4 & 65.4 \\
\noalign{\vskip 2pt}
~~+ Visual Sensitivity & 41.5 & 68.8 & 67.2 & \textbf{67.8} & 64.8 & 73.9 & 64.0 & 78.5 & \textbf{79.2} & 78.2 & 61.4 & 74.3 & 68.1 \\
~~+ Visual Robustness & 40.9 & 67.2 & 66.6 & 67.0 & 63.2 & 73.0 & 63.0 & 76.2 & 76.5 & \textbf{80.2} & 59.1 & 73.0 & 67.0 \\
\noalign{\vskip 2pt}
\rowcolor{gainbg}
\textbf{DVRP$_G$ (Full)} & \textbf{42.3} & \textbf{71.1} & \textbf{68.1} & 67.4 & \textbf{66.7} & \textbf{75.6} & \textbf{65.2} & \textbf{81.7} & 78.3 & 79.9 & \textbf{65.5} & \textbf{76.4} & \textbf{69.7} \\
\bottomrule
\end{tabular}%
}
\caption{Ablation study of \textbf{Visual Sensitivity} (via masking) and \textbf{Visual Robustness} (via noise injection) on the Qwen2.5-VL-7B backbone. The \textbf{Base Model} (gray) represents the original pre-trained weights, while the \textbf{Baseline} utilizes the standard GRPO algorithm. While specific constraints (e.g., masking on PathVQA or noise on VQA-RAD) may yield marginal gains on individual datasets, the full \textbf{DVRP$_G$} framework (green) achieves the best overall performance, demonstrating the synergy of the combined objectives.}
\label{tab:ablation_study}
\end{table*}

\subsection{Main Experiments}
\subsubsection{RLVR Experiments}
We benchmark DVRP against representative RLVR baselines~\cite{noisyrollout} (Table~\ref{tab:main_results}). On the 7B scale, our approach consistently establishes new state-of-the-art results. Implementing DVRP on GRPO (DVRP$_G$) yields an overall accuracy of 69.7\%, representing a \textbf{40.5\%} relative improvement over the base model and surpassing the strongest baseline PAPO by 2.5 points. In the medical domain, DVRP$_G$ reaches 76.4\% accuracy with a relative gain of \textbf{28.4\%}, demonstrating differential visual constraints effectively enhance robustness in specialized reasoning tasks. For further case studies and reasoning consistency evaluation, please refer to Section~\ref{sec:algorithm_comparison}.

The performance gain is even more pronounced on the smaller 3B backbone. DVRP$_D$ boosts the overall accuracy from 37.4\% to 62.7\%, achieving a remarkable \textbf{67.6\%} relative improvement. Crucially, our 3B model attains 73.3\% on medical benchmarks, which surpasses the 70.4\% accuracy of the significantly larger 7B GRPO baseline. These results validate that DVRP effectively compensates for limited model capacity by maximizing the utilization of visual evidence.

\subsubsection{Domain Foundation Model Experiments}
We evaluate DVRP against a wide spectrum of proprietary, general-purpose, and domain-specific foundation models across medical and mathematical benchmarks. In the medical domain, as detailed in Table~\ref{tab:medical_results}, our 7B model establishes new state-of-the-art performance on in-domain tasks, consistently outperforming specialized baselines like CAPO-7B~\cite{capo} and HuatuoGPT-Vision-34B~\cite{huatuo}. Notably, on PathVQA, DVRP$_D$ achieves \textbf{79.7\%} accuracy, surpassing the proprietary GPT-4o score of 75.9\% and rivaling complex agentic systems like MMedAgent-RL without requiring external tools or trajectory data~\cite{aflow}.

Turning to the mathematical domain in Table~\ref{tab:math_results}, DVRP demonstrates superior generalization capabilities. Our 7B model attains an average accuracy of \textbf{65.2\%}, significantly outperforming the 72B-parameter Qwen2.5-VL at 57.9\% and GPT-4o at 57.5\%. It also establishes a clear advantage over math-specific baselines such as RACRO-7B~\cite{racro-7b} and MM-Eureka~\cite{mm-eureka}, which score 61.7\% and 59.0\% respectively. These results indicate that by explicitly enforcing differential visual sensitivity and robustness, DVRP ensures consistent generalization across diverse datasets and model scales. Crucially, this approach empowers parameter-efficient models to rival or even exceed the reasoning capabilities of significantly larger counterparts and more sophisticated agentic reasoning systems~\cite{mmedagents}.

\subsection{Ablation Experiments}
To investigate the individual contributions of the proposed components, we conduct an ablation study on the Qwen2.5-VL-7B backbone using GRPO as the baseline (Table~\ref{tab:ablation_study}).
The introduction of \textit{Visual Sensitivity} yields a substantial improvement over the baseline, confirming that penalizing blind reasoning is critical for enforcing genuine visual grounding.
Similarly, the \textit{Visual Robustness} term independently enhances performance by fostering stability against perturbations.
While individual components may occasionally outperform the unified model on specific tasks, the full DVRP framework achieves the best overall performance.
This demonstrates that sensitivity and robustness constraints are complementary, effectively synergizing to maximize reasoning reliability across diverse domains.

Furthermore, we explore the impact of perturbation intensity across domains.
Our experiments reveal a domain-specific dichotomy: general multimodal reasoning (e.g., Math) benefits from aggressive perturbations ($P_{mask}=0.6$, $T_{init}=500$) to enforce structural dependency, whereas medical reasoning requires milder regularization ($P_{mask}=0.2$, $T_{init}=100$) to preserve fine-grained pathological features.
We provide a detailed discussion of these hyperparameter sensitivities in Appendix~\ref{sec:pertubation_ablation}.

\section{Conclusion}
In this work, we address the critical perception-reasoning decoupling in current multimodal RLVR paradigms by proposing \textbf{Thinking with Deltas}. Driven by the \textbf{Differential Visual Reasoning Policy}, our framework leverages self-supervised \textbf{visual triplets} to introduce intrinsic supervision, compelling models to strictly align their reasoning with the presence and stability of visual evidence. This approach natively enhances perception without relying on external dependencies.Extensive experiments across mathematical and medical domains demonstrate that DVRP effectively bridges the perception-reasoning decoupling and empowers parameter-efficient models (e.g., 3B and 7B) to achieve state-of-the-art performance, often rivaling significantly larger commercial baselines.

\section*{Limitations}

Despite the robust performance and generalization capabilities our method demonstrates across diverse domains and comparative settings, several limitations remain.
First, our empirical evaluation is currently confined to models with 3B and 7B parameters. Due to computational resource constraints, we have not yet scaled the proposed framework to larger foundation models (e.g., 70B+ parameters) or verified its efficacy across a broader spectrum of architectural backbones. Consequently, the scalability of differential visual constraints on massive-scale models and their transferability to different model families remain to be fully characterized.
Second, our current approach focuses on intrinsic policy optimization and has not yet explored integration with agentic systems. Synergizing DVRP with multi-agent frameworks or interactive tool-use pipelines to achieve more sophisticated reasoning strictly grounded on visual evidence remains an unexplored frontier.
We hope this work inspires future research into differential constraints as a minimalist yet powerful paradigm for building trustworthy and visually grounded multimodal systems.

\bibliography{acl_latex}

\appendix
\section{Appendix}
\label{sec:appendix}

This appendix provides implementation details, additional quantitative ablations, and extensive qualitative examples to support the findings in the main paper. The organization is as follows:

\begin{itemize}
    \setlength{\itemsep}{0.5em}
    \item \textbf{\S\ref{sec:implementation_details}Implementation Details}. We provide a detailed breakdown of our experimental setup, including:
    \begin{itemize}
        \item \textbf{\S\ref{sec:evaluation} Evaluation Details}: Detailed protocols for Mathematical and Medical benchmarks.
        \item \textbf{\S\ref{sec:hyperparameter} Hyperparameter}: Comprehensive lists of training hyperparameters and DVRP coefficients.
        \item \textbf{\S\ref{sec:prompt} Prompt Template}: The standardized system prompt and reasoning template used across all experiments.
        \item \textbf{\S\ref{sec:noise} Noise Scheduling}: We detail the variance-preserving diffusion formulation and the Sigmoid annealing schedule employed to dynamically modulate noise intensity, implementing a curriculum learning strategy for visual robustness.
    \end{itemize}
    
    \item \textbf{\S\ref{sec:pertubation_ablation} Ablation on Perturbation Parameters}. We present the sensitivity analysis for masking ratios ($P_{mask}$) and noise steps ($T_{init}$).
    
    \item \textbf{\S\ref{sec:robustness_analysis} Visual Robustness and Sensitivity Analysis}. We analyze the contrasting behaviors of MLLMs under diffusion noise versus semantic masking.
    
    \item \textbf{\S\ref{sec:rlvr_visual_dependency} RLVR Visual Dependency Experiments}. We detail the blind experiments (Text-Only and Blank Image) used to verify visual grounding.
    
    \item \textbf{\S\ref{sec:qualitative} Qualitative Analysis of Visual Dependency}. We provide case studies demonstrating linguistic shortcuts in baseline methods under blind settings.
    
    \item \textbf{\S\ref{sec:algorithm_comparison} Qualitative Comparison of RLVR Algorithms}. We offer a comparative visualization of reasoning trajectories between GRPO, DAPO, and our DVRP-D.
\end{itemize}

\section{Implementation Details}
\label{sec:implementation_details}
\subsection{Evaluation Details}
\label{sec:evaluation}

\vspace{0.5em}
\noindent \textbf{Evaluation Protocols.} To comprehensively assess model performance, we conducted evaluations across both general (mathematical) and medical domains, distinguishing between in-domain and out-of-domain (OOD) settings.
\begin{itemize}
    \item \textbf{Mathematical Evaluation:} We employed Geo3k~\cite{geo3k}, Vista~\cite{mathvista}, WeMath~\cite{wemath}, MVerse~\cite{mathverse}, MVerse-V~\cite{mathverse}, and MMKI2~\cite{mmk12} as OOD benchmarks to test generalization capabilities. Furthermore, we utilized \texttt{mathruler.grader} to facilitate precise evaluation.
    \item \textbf{Medical Evaluation:} We utilized the test splits of Slake~\cite{slake}, PathVQA~\cite{pathvqa}, RadVQA~\cite{radvqa}, and PMC-VQA~\cite{pmcvqa} for in-domain evaluation. Furthermore, MedXpertQA~\cite{medxpertqa} and MMMU-Med~\cite{mmmu} were employed to assess OOD performance.
\end{itemize}
To ensure the statistical reliability of our results, we report the Average Accuracy over 8 runs (\textbf{AVG@8 Acc}). For inference, we deployed the vLLM engine to accelerate generation~\cite{vllm}. For fair comparison, all models utilized a unified system prompt (reasoning template) to elicit chain-of-thought reasoning, with the temperature set to 1.0 and top-p to 0.9. The specific templates used are detailed in Appendix~\ref{sec:prompt}.

\vspace{0.5em}
\noindent \textbf{Baselines.} To conduct a comprehensive comparative analysis, we evaluate our proposed method against baselines categorized into two distinct dimensions: RL-based optimization methods and foundational MLLMs.

\noindent \textit{RL-based Methods.} We benchmark our approach against state-of-the-art RL strategies designed for reasoning or visual alignment, including \textbf{GRPO}~\cite{shao2024deepseekmathpushinglimitsmathematical}, \textbf{DAPO}~\cite{dapo}, \textbf{PAPO}~\cite{papo}, and \textbf{NoiseRollout}~\cite{noisyrollout}. To ensure a fair comparison, we re-trained these RL methods using the identical training set and evaluation suite.

\noindent \textit{Foundation Models.} We compare against a diverse spectrum of baselines ranging from proprietary commercial systems to specialized open-source models:
\begin{itemize}
    \item \textit{Medical-Specific Models:} We evaluate performant open-source models tailored for the biomedical domain, such as \textbf{HuatuoGPT-Vision-34B}~\cite{huatuo}.
    \item \textit{Math-Specific Models:} For mathematical reasoning tasks, we include domain-expert models like \textbf{MM-Eureka}~\cite{mm-eureka} and \textbf{ThinkLite-7B}~\cite{thinklite}.
    \item \textit{General-Purpose Models:} We also include widely adopted general open-source benchmarks, such as \textbf{Qwen-VL}~\cite{bai2023qwen} and \textbf{LLaVA-v1.6}~\cite{llava-v1.6}.
\end{itemize}

\subsection{Hyperparameter}
\label{sec:hyperparameter}
To ensure reproducibility, we provide a comprehensive overview of the training and evaluation hyperparameters in Table~\ref{tab:hyperparams} and Table~\ref{tab:hyperparams_dapo}. 
All models are optimized using AdamW with a constant learning rate of $1\text{e-}6$ and a global batch size of 128. 
Consistent with recent reinforcement learning practices, we remove the KL penalty relative to the reference model to facilitate broader policy exploration~\cite{noisyrollout, dapo}.

Crucially, our \textbf{DVRP} framework introduces three auxiliary objectives regulated by specific coefficients: Visual Robustness ($\lambda_{rob}=0.01$), Visual Sensitivity ($\lambda_{nec}=0.01$), and Entropy Regularization ($\lambda_{ent}=0.05$).
The above loss weights are consistent with previous works~\cite{papo, noisyrollout}. To address the varying information density of visual modalities, we employ domain-adaptive perturbation strategies.
As detailed in Table~\ref{tab:hyperparams}, we apply a higher mask probability ($P_{mask}=0.6$) and longer noise injection steps ($T_{init}=500$) for mathematical reasoning, while adopting milder perturbations ($P_{mask}=0.2, T_{init}=100$) for medical tasks to preserve fine-grained pathological details.

\begin{table}[t!]
\centering
\resizebox{\linewidth}{!}{
\begin{tabular}{ll}
\toprule
\textbf{Hyperparameter} & \textbf{Value} \\
\midrule
\textit{General Training} & \\
Optimizer & AdamW (BF16) \\
Learning Rate & 1e-6 \\
LR Schedule & Constant \\
Total Epochs & 3 \\
\midrule
\textit{RL Process (GRPO)} & \\
Global Batch Size & 128 \\
Rollout Batch Size & 384 \\
Group Size ($G$) & 5 \\
KL Penalty (Ref Model) & None (Disabled) \\
Reward Signal & Accuracy \\
\midrule
\textit{DVRP Objectives} & \\
Visual Robustness Loss ($\lambda_{rob}$) & 0.01 \\
Visual Sensitivity Loss ($\lambda_{nec}$) & 0.01 \\
Entropy Regularization ($\lambda_{ent}$) & 0.05 \\
\midrule
\textit{Visual Perturbation} & \textbf{Math} \quad / \quad \textbf{Medical} \\
Mask Probability ($P_{mask}$) & 0.6 \quad / \quad 0.2 \\
Patch Size ($P_{mask}$) & 14 \quad / \quad 14 \\
Noise Steps ($T_{init}$) & 500 \quad / \quad 100 \\
\midrule
\textit{Inference} & \\
Temperature & 1.0 \\
Top-p & 0.99 \\
Max New Tokens & 2048 \\
\midrule
\textit{Resources \& Efficiency} & \\
Compute & 4$\times$ NVIDIA A800 \\
Step Time & $\sim$1000--2000s \\
\bottomrule
\end{tabular}
}
\caption{Key hyperparameters for training and evaluation. DVRP parameters vary by domain (Math vs. Medical). Notations correspond to Eq. (4).}
\label{tab:hyperparams}
\end{table}

\begin{table}[t!]
\centering
\resizebox{\linewidth}{!}{
\begin{tabular}{ll}
\toprule
\textbf{Hyperparameter} & \textbf{Value} \\
\midrule
\textit{General Training} & \\
Optimizer & AdamW (BF16) \\
Learning Rate & 1e-6 \\
LR Schedule & Constant \\
Total Epochs & 3 \\
\midrule
\textit{RL Process (DAPO)} & \\
Global Batch Size & 128 \\
Rollout Batch Size & 384 \\
Mini-Rollout Batch Size & 128 \\
Group Size ($G$) & 5 \\
KL Penalty (Ref Model) & None (Disabled) \\
Reward Signal & Accuracy \\
\midrule
\textit{DVRP Objectives} & \\
Visual Robustness Loss ($\lambda_{rob}$) & 0.01 \\
Visual Sensitivity Loss ($\lambda_{nec}$) & 0.01 \\
Entropy Regularization ($\lambda_{ent}$) & 0.05 \\
\midrule
\textit{Visual Perturbation} & \textbf{Math} \quad / \quad \textbf{Medical} \\
Mask Probability ($P_{mask}$) & 0.6 \quad / \quad 0.2 \\
Patch Size ($P_{mask}$) & 14 \quad / \quad 14 \\
Noise Steps ($T_{init}$) & 500 \quad / \quad 100 \\
\midrule
\textit{Inference} & \\
Temperature & 1.0 \\
Top-p & 0.99 \\
Max New Tokens & 2048 \\
\midrule
\textit{Resources \& Efficiency} & \\
Compute & 4$\times$ NVIDIA A800 \\
Step Time & $\sim$1000--2000s \\
\bottomrule
\end{tabular}
}
\caption{Key hyperparameters for training and evaluation based on the DAPO + DVRP configuration.}
\label{tab:hyperparams_dapo}
\end{table}

\begin{table*}[t!]
\centering
\resizebox{\textwidth}{!}{%
\begin{tabular}{l | cccccc c | cccc c | c}
\toprule
\multirow{2}{*}{\textbf{Method / Setting}} & 
\multicolumn{7}{c|}{\cellcolor{generalbg}\textbf{General Multimodal Reasoning}} & 
\multicolumn{5}{c|}{\cellcolor{medicalbg}\textbf{Medical Multimodal Reasoning}} & 
\cellcolor{overallbg}\textbf{Overall} \\
\cmidrule{2-14}
 & \textbf{Geo3k} & \textbf{Vista} & \textbf{WeMath} & \textbf{MVerse} & \textbf{MVerse-V} & \textbf{MMKI2} & \textbf{AVG} & \textbf{Slake} & \textbf{Path} & \textbf{Rad} & \textbf{PMC} & \textbf{AVG} & \textbf{AVG} \\
\midrule

\multicolumn{14}{c}{\textit{Qwen2.5-VL-7B Backbone}} \\
\midrule
\rowcolor{graybg}
Base Model (Original) & 33.8 & 55.9 & 41.8 & 45.6 & 36.9 & 43.7 & 43.0 & 63.7 & 60.6 & 61.3 & 52.2 & 59.5 & 49.6 \\
Baseline (GRPO)       & 40.2 & 65.5 & 66.1 & 66.5 & 61.7 & 72.1 & 62.0 & 74.7 & 75.7 & 74.9 & 56.2 & 70.4 & 65.4 \\
\midrule

\multicolumn{14}{l}{\textit{\textbf{Ablation I: Sensitivity to Masking Ratio ($P_{mask}$)}} \quad \textcolor{gray}{\footnotesize{(Math favors high mask, Medical favors low mask)}}} \\
\midrule
+ Masking ($P=0.2$) & 41.0 & 68.2 & 66.8 & 66.9 & 63.5 & 73.5 & 63.3 & 78.5 & \textbf{79.2} & 78.2 & 61.4 & \textbf{74.3} & 67.7 \\
+ Masking ($P=0.4$) & 41.8 & 69.5 & 67.5 & 67.1 & 65.1 & 74.8 & 64.3 & 76.8 & 78.1 & 77.4 & 60.2 & 73.1 & 67.8 \\
+ Masking ($P=0.6$) & \textbf{42.3} & \textbf{71.1} & \textbf{68.1} & \textbf{67.4} & \textbf{66.7} & \textbf{75.6} & \textbf{65.2} & 74.5 & 76.5 & 75.8 & 58.9 & 71.4 & 67.7 \\
\midrule

\multicolumn{14}{l}{\textit{\textbf{Ablation II: Robustness to Noise Injection ($T_{init}$)}} \quad \textcolor{gray}{\footnotesize{(Math needs strong noise, Medical needs weak noise)}}} \\
\midrule
+ Noise ($T=100$) & 41.2 & 68.5 & 66.9 & 67.0 & 64.2 & 73.9 & 63.6 & 76.2 & 76.5 & \textbf{80.2} & 59.1 & \textbf{73.0} & 67.4 \\
+ Noise ($T=300$) & 41.9 & 70.1 & 67.6 & 67.2 & 65.8 & 74.9 & 64.6 & 75.1 & 75.8 & 78.4 & 57.8 & 71.8 & 67.5 \\
+ Noise ($T=500$) & \textbf{42.3} & \textbf{71.1} & \textbf{68.1} & \textbf{67.4} & \textbf{66.7} & \textbf{75.6} & \textbf{65.2} & 73.5 & 74.2 & 76.1 & 56.5 & 70.1 & 67.2 \\
\midrule

\rowcolor{gainbg}
\textbf{DVRP$_G$ (Optimal)} & \textbf{42.3} & \textbf{71.1} & \textbf{68.1} & 67.4 & \textbf{66.7} & \textbf{75.6} & \textbf{65.2} & \textbf{81.7} & 78.3 & 79.9 & \textbf{65.5} & \textbf{76.4} & \textbf{69.7} \\
\bottomrule
\end{tabular}%
}
\caption{Comprehensive ablation study of the DVRP framework on Qwen2.5-VL-7B. We systematically vary the \textbf{Masking Ratio} ($P_{mask} \in \{0.2, 0.4, 0.6\}$) and \textbf{Noise Steps} ($T_{init} \in \{100, 300, 500\}$). The results highlight a domain-specific dichotomy: \textbf{General/Math} tasks benefit from aggressive perturbations ($P=0.6, T=500$) to enforce structural reasoning, whereas \textbf{Medical} tasks require milder regularization ($P=0.2, T=100$) to preserve fine-grained pathological features. The final \textbf{DVRP$_G$} model employs these domain-optimal settings.}
\label{tab:full_ablation}
\end{table*}

\begin{table}[t!]
\centering
\resizebox{\linewidth}{!}{%
\begin{tabular}{l | cccc c}
\toprule
\multirow{2}{*}{\textbf{Method / Setting}} & 
\multicolumn{5}{c}{\cellcolor{medicalbg}\textbf{Medical Multimodal Reasoning}} \\
\cmidrule{2-6}
 & \textbf{Slake} & \textbf{Path} & \textbf{Rad} & \textbf{PMC} & \textbf{AVG} \\
\midrule

\rowcolor{graybg}
Base Model (Original) & 63.7 & 60.6 & 61.3 & 52.2 & 59.5 \\
Baseline (GRPO)       & 74.7 & 75.7 & 74.9 & 56.2 & 70.4 \\
\midrule

\multicolumn{6}{l}{\textit{\textbf{Comparison of Noise Types}} \quad \textcolor{gray}{\footnotesize{(Fixed scheduling, varying structure)}}} \\
\midrule
+ Gaussian Noise ($\sigma=0.1$) & 75.2 & 76.0 & 76.5 & 57.0 & 71.2 \\
\rowcolor{gainbg}
+ Diffusion Noise ($T=100$)     & \textbf{76.2} & \textbf{76.5} & \textbf{80.2} & \textbf{59.1} & \textbf{73.0} \\
\bottomrule
\end{tabular}%
}
\caption{Ablation study comparing noise injection strategies in the Medical domain. Under the DVRP framework, structured \textbf{Diffusion Noise} ($T=100$) outperforms unstructured \textbf{Gaussian Noise} ($\sigma=0.1$) by \textbf{+1.8\%}, demonstrating that preserving semantic structure during perturbation is critical for medical visual reasoning.}
\label{tab:medical_noise_ablation}
\end{table}

\subsection{Prompt Template}
\label{sec:prompt}
For the prompt selection, we employ a standardized template across all training and evaluation stages to elicit stable Chain-of-Thought (CoT) reasoning. 
As illustrated in Figure~\ref{fig:prompt_template}, the template explicitly instructs the model to generate an internal monologue within \texttt{<think>} tags before producing the final answer, ensuring the output format remains consistent for automated parsing and reward calculation.

\subsection{Noise Scheduling}
\label{sec:noise}

We adopt a variance-preserving (VP) diffusion process to construct the incremental view $I_{\text{noise}}$, simulating environmental instability while maintaining semantic consistency. Given an original image $I$ and a noise intensity coefficient $\beta \in [0, 1]$, the perturbed input is formulated as:
\begin{equation}
    I_{\text{noise}} = \sqrt{1 - \beta} \cdot I + \sqrt{\beta} \cdot \epsilon, \quad \epsilon \sim \mathcal{N}(0, \mathbf{I})
\end{equation}
where $\epsilon$ represents standard Gaussian noise. The coefficient $\beta$ determines the signal-to-noise ratio; a higher $\beta$ introduces stronger perturbation.

To balance structural robustness learning with convergence stability, we introduce a curriculum learning strategy via a Sigmoid decay schedule. Specifically, the noise intensity $\beta$ is dynamically annealed based on the training progress. Let $k$ be the current training step and $K$ be the total steps. We compute the diffusion timestep $t$ as:
\begin{equation}
    t(k) = T_{\text{init}} \cdot \sigma\left( \gamma \cdot \left(0.5 - \frac{k}{K}\right) \right),
\end{equation}
where $T_{\text{init}}$ is the initial noise step (e.g., 500), $\sigma(\cdot)$ is the sigmoid function, and $\gamma$ is a scaling factor controlling the decay steepness (set to 10 in our experiments). The noise intensity is then derived as $\beta_k = t(k) / T_{\text{max}}$. This schedule imposes high-variance perturbations in the early stages to enforce broad structural invariance, while gradually annealing the noise to zero to refine fine-grained visual consistency.

\section{Ablation on Perturbation Parameters}
\label{sec:pertubation_ablation}

We investigate the sensitivity of the DVRP framework to the intensity of visual perturbations. 
We perform ablations on the Qwen2.5-VL-7B~\cite{qwen2.5} backbone by varying the masking ratio $P_{mask} \in \{0.2, 0.4, 0.6\}$ and the noise injection steps $T_{init} \in \{100, 300, 500\}$. 
The results are detailed in Table~\ref{tab:full_ablation}.

\paragraph{Masking Ratio.}
We observe a distinct divergence in optimal masking ratios between domains.
For \textbf{General Multimodal Reasoning} (e.g., Math), performance improves consistently as the masking ratio increases, peaking at $P_{mask}=0.6$. 
This suggests that geometric diagrams and natural images contain high spatial redundancy; aggressive masking effectively forces the model to learn structural dependencies rather than relying on local textures.
In contrast, \textbf{Medical Multimodal Reasoning} favors a conservative ratio ($P_{mask}=0.2$). 
Performance drops significantly (from 74.3\% to 71.4\%) when the ratio is increased to 0.6. 
This indicates that medical diagnosis relies heavily on fine-grained visual details (e.g., small lesions or boundaries), which are easily destroyed by aggressive occlusion.

\paragraph{Noise Injection Level.}
A similar trend is observed for noise robustness.
We implement noise injection via a diffusion process, utilizing a sigmoid schedule to modulate the noise intensity.
In the general domain, the model benefits from stronger consistency regularization, achieving peak accuracy with higher diffusion timesteps ($T_{init}=500$).
Conversely, the medical domain is sensitive to high-intensity diffusion perturbations.
Performance degrades as $T_{init}$ increases, with the optimal setting found at a lower timestep ($T_{init}=100$).
We hypothesize that excessive diffusion noise in medical images risks corrupting subtle pathological features or mimicking sensor artifacts, thereby confusing the reasoning policy.

Based on these observations, our final \textbf{DVRP$_G$} model adopts domain-specific settings: we apply stronger perturbations ($P=0.6, T=500$) for math tasks to encourage robustness, and milder perturbations ($P=0.2, T=100$) for medical tasks to preserve visual fidelity.

\paragraph{Impact of Noise Structure.}
Table~\ref{tab:medical_noise_ablation} further scrutinizes the efficacy of structured \textit{Diffusion Noise} ($T=100$) versus unstructured \textit{Gaussian Noise} ($\sigma=0.1$) in the medical domain ($P=0.6$).
The results demonstrate that Diffusion Noise consistently outperforms Gaussian Noise, achieving an average gain of \textbf{+1.8\%}.
This suggests that structured perturbations effectively regularize the policy while preserving essential semantic integrity, whereas unstructured Gaussian noise risks degrading the pixel-level fidelity required for fine-grained medical diagnostics.

\begin{figure*}[t]
    \centering
    \definecolor{promptbg}{RGB}{240, 248, 255} 
    \definecolor{framecolor}{RGB}{0, 0, 0} 

    \begin{tcolorbox}[
        colback=promptbg,      
        colframe=framecolor,   
        coltitle=black,        
        title=\textbf{Reasoning Template}, 
        fonttitle=\large,      
        boxrule=0.8pt,         
        arc=3mm,               
        left=6pt, right=6pt, top=6pt, bottom=6pt, 
        sharp corners=south,   
    ]
        \textbf{SYSTEM:}\\
        You are a helpful assistant.
        
        \vspace{1em} 
        
        \textbf{USER:}\\
        \textcolor{red}{\{question\}} 
        
        \vspace{1em}
        
        You first think through the reasoning process as an internal monologue, enclosed within $<$think$>$ $<$/think$>$ tags. Then, provide your final answer enclosed within $\backslash$boxed\{\}.
    \end{tcolorbox}
    \caption{The standardized prompt template employed across all training and evaluation phases. To ensure consistent reasoning behaviors and facilitate automated answer extraction, we explicitly instruct the model to encapsulate its chain-of-thought within \texttt{<think>} tags and place the final result inside a \texttt{$\backslash$boxed\{\}} command.}
    \label{fig:prompt_template}
\end{figure*}

\begin{figure}[t]
    \centering
    \includegraphics[width=\linewidth]{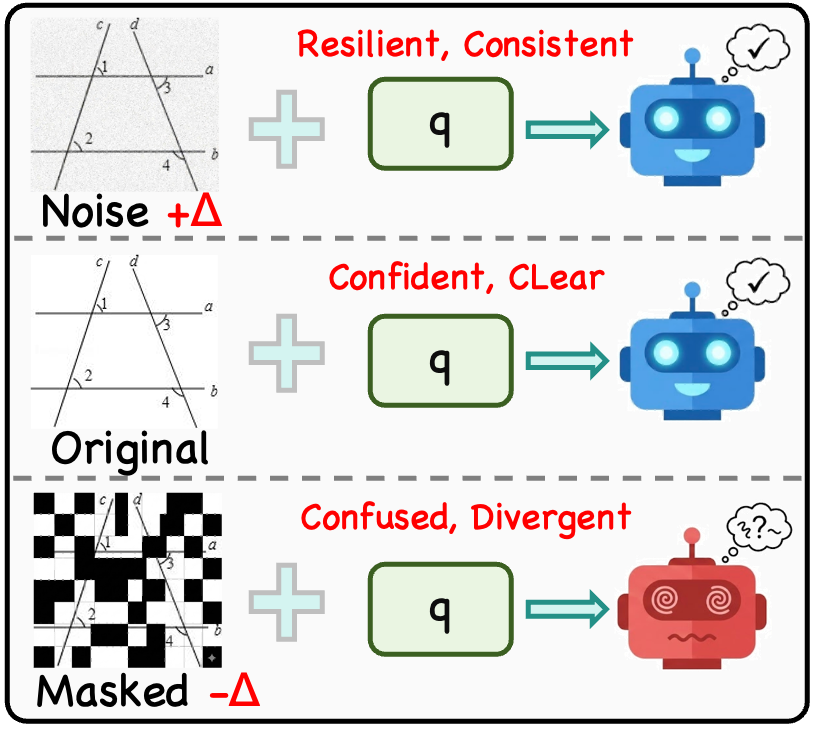} 
    \caption{\textbf{Visual Dependency Analysis.} 
    Divergence in reasoning paths. Current MLLMs show minimal sensitivity to masked vs.\ perturbed inputs (consistent in top rows), but become confused when visual semantics are masked (divergent in bottom row), indicating a lack of visual grounding.}
    \label{fig:visual_dependency}
\end{figure}

\begin{table*}[t!]
\centering
\resizebox{\textwidth}{!}{%
\begin{tabular}{l | cccccc c | cccc c}
\toprule
\multirow{2}{*}{\textbf{Setting}} & 
\multicolumn{7}{c|}{\cellcolor{generalbg}\textbf{General Multimodal Reasoning}} & 
\multicolumn{5}{c}{\cellcolor{medicalbg}\textbf{Medical Multimodal Reasoning}} \\
\cmidrule{2-13}
 & \textbf{Geo3k} & \textbf{Vista} & \textbf{WeMath} & \textbf{MVerse} & \textbf{MVerse-V} & \textbf{MMK12} & \textbf{AVG} & 
 \textbf{Slake} & \textbf{Path} & \textbf{Rad} & \textbf{PMC} & \textbf{AVG} \\
\midrule

\multicolumn{13}{c}{\textit{\textbf{Qwen2.5-VL-3B Backbone}}} \\
\midrule
\rowcolor{graybg}
Original & 20.6 & 40.6 & 23.9 & 30.9 & 28.2 & 34.8 & 29.8 & 48.7 & 59.2 & 40.3 & 46.8 & 48.8 \\
\addlinespace[3pt] 

Noise & \p{22.6}{2.0} & \p{41.4}{0.8} & \p{26.5}{2.6} & \m{30.8}{-0.1} & \m{26.2}{-2.0} & \p{35.1}{0.3} & \p{30.4}{0.6} & \m{48.1}{-0.6} & \m{58.5}{-0.7} & \p{40.6}{0.3} & \m{46.2}{-0.6} & \m{48.4}{-0.4} \\
\addlinespace[2pt]

Masked & \m{6.2}{-14.4} & \m{24.2}{-16.4} & \m{17.8}{-6.1} & \m{21.5}{-9.4} & \m{16.5}{-11.7} & \m{31.6}{-3.2} & \m{19.6}{-10.2} & \m{31.5}{-17.2} & \m{37.8}{-21.4} & \m{26.2}{-14.1} & \m{29.4}{-17.4} & \m{31.2}{-17.6} \\

\midrule
\multicolumn{13}{c}{\textit{\textbf{Qwen2.5-VL-7B Backbone}}} \\
\midrule
\rowcolor{graybg}
Original & 33.8 & 55.9 & 41.8 & 45.6 & 36.9 & 43.7 & 42.9 & 63.7 & 60.6 & 61.3 & 52.2 & 59.5 \\
\addlinespace[3pt]

Noise & \m{30.5}{-3.3} & \p{56.3}{0.4} & \m{41.6}{-0.2} & \m{44.5}{-1.1} & \p{38.5}{1.6} & \p{44.6}{0.9} & \m{42.7}{-0.2} & \m{63.5}{-0.2} & \m{59.8}{-0.8} & \p{61.5}{0.2} & \m{51.9}{-0.3} & \m{59.2}{-0.3} \\
\addlinespace[2pt]

Masked & \m{8.0}{-25.8} & \m{32.4}{-23.5} & \m{30.5}{-11.3} & \m{31.2}{-14.4} & \m{23.2}{-13.7} & \m{39.6}{-4.1} & \m{27.5}{-15.4} & \m{42.1}{-21.6} & \m{38.4}{-22.2} & \m{40.5}{-20.8} & \m{33.2}{-19.0} & \m{38.6}{-20.9} \\

\bottomrule
\end{tabular}%
}
\caption{Robustness evaluation under visual perturbations. Results are presented as \textbf{Accuracy\textsubscript{Difference}}. The difference ($\Delta$) represents the gap compared to the \textbf{Original} setting. \textcolor{plusColor}{Red subscripts} indicate performance gains ($+\Delta$), while \textcolor{minusColor}{Blue subscripts} indicate drops ($-\Delta$).}
\label{tab:robustness_full}
\end{table*}

\section{Visual Robustness and Sensitivity Analysis}
\label{sec:robustness_analysis}
To investigate the reliance of MLLMs on visual fidelity and their resilience to perturbations, we conduct a comprehensive robustness analysis on Qwen2.5-VL benchmarks~\cite{qwen2.5} across general and medical domains.
We introduce two distinct types of visual perturbations: Diffusion Noise, which represents high-frequency corruption, and Patch Masking, which represents semantic information loss.
As visualized in Figure~\ref{fig:visual_dependency}, these perturbations induce contrasting reasoning behaviors: the model remains resilient and consistent under noise ($+\Delta$) but becomes confused and divergent when visual semantics are masked ($-\Delta$).
This qualitative observation is quantitatively confirmed by the results in Table~\ref{tab:robustness_full}, serving as the empirical foundation for our proposed method.

\paragraph{Robustness to Diffusion Noise.}
As shown in the Noise rows of Table~\ref{tab:robustness_full}, MLLMs exhibit strong stability against diffusive perturbations.
We implement noise injection via a diffusion scheduler controlled by a sigmoid function.
Despite these perturbations, the Qwen2.5-VL-3B model maintains performance and even achieves a slight average gain of \textbf{+0.6\%} in general reasoning tasks.
Similarly, the 7B model shows negligible performance fluctuations ($\Delta < 0.3\%$) across most benchmarks.
This indicates that current MLLMs possess inherent resilience to the high-frequency jitter introduced by the diffusion scheduler, likely due to the robust feature extraction capabilities of the vision encoder.

\textbf{Insight I:} This robustness validates the feasibility of our \textit{Noise Consistency} objective.
Since the model demonstrates invariance to diffusion noise, explicitly enforcing output consistency between clean and noisy views acts as a safe regularization term.
This stabilizes the policy without degrading representation quality.

\paragraph{Sensitivity to Semantic Masking.}
Conversely, the models demonstrate high sensitivity to patch masking.
When a portion of the visual input is occluded, performance degrades sharply.
Notably, the 7B model suffers a severe average drop of \textbf{$-$20.9\%} in the medical domain and \textbf{$-$15.4\%} in general tasks.
This degradation is significantly more pronounced than that caused by noise, indicating that reasoning depends heavily on specific visual patches rather than language priors alone.

\textbf{Insight II:} This sensitivity confirms that visual tokens are critical for correct reasoning.
It motivates employing masking as a challenging view in the reinforcement learning loop.
Exposing the policy to masked inputs forces the model to maximize information utilization from the remaining visible patches, thereby reducing hallucination and enhancing the grounding of the reasoning process.

\paragraph{Summary.}
MLLMs process these two perturbations through fundamentally different mechanisms.
Diffusion noise acts as a low-level surface artifact to which pre-trained visual encoders exhibit strong invariance.
In contrast, patch masking acts as a high-level semantic disruption that breaks the visual continuity required for logical deduction.
This disparity highlights a critical insight: while visual representations are texturally robust, reasoning policies remain fragile to structural incompleteness.
\textit{This observation directly underpins our methodology, where we leverage diffusion noise for consistency regularization and masking for hardness-aware policy learning.}

\section{RLVR Visual Dependency Experiments}
\label{sec:rlvr_visual_dependency}
To rigorously investigate whether current RLVR methods genuinely leverage visual information or merely exploit linguistic priors, we conducted a series of \textit{blind experiments}.
We maintained the identical training setup as the main experiments, training policies on the VIRL-39K~\cite{vl-rethinker} dataset using the standard hyperparameters for both GRPO~\cite{shao2024deepseekmathpushinglimitsmathematical} and DAPO~\cite{capo}.
We also presented case study comparisons in Section~\ref{sec:qualitative}.
During evaluation, we introduced two modality-blind settings to sever the visual semantic link:

\begin{itemize}
    \item \textbf{Blank Image (B/W):} The original visual input is randomly replaced with a solid black or white image. This removes all visual semantics while preserving the multimodal input structure.
    \item \textbf{Text Only:} The visual tokens are entirely discarded, forcing the model to rely solely on the textual query and its internal parametric knowledge.
\end{itemize}

\begin{table*}[t!]
\centering
\resizebox{\textwidth}{!}{%
\begin{tabular}{l | cccccc | c}
\toprule
\multirow{2}{*}{\textbf{Method \& Setting}} & 
\multicolumn{7}{c}{\cellcolor{basegray!20}\textbf{General Multimodal Reasoning Benchmarks}} \\
\cmidrule{2-8}
 & \textbf{Geo3k} & \textbf{Vista} & \textbf{WeMath} & \textbf{MVerse} & \textbf{MVer-V} & \textbf{MMK12} & \textbf{AVG} \\
\midrule

\rowcolor{basegray}
\textbf{Base Model} (Qwen2.5-VL-7B) & 33.8 & 55.9 & 41.8 & 45.6 & 36.9 & 43.7 & 43.0 \\
\addlinespace[5pt]

\rowcolor{methodblue}
\textbf{GRPO Policy} (Original) & 40.2 & 65.5 & 66.1 & 66.5 & 61.7 & 72.1 & 62.0 \\
\addlinespace[2pt]
\hspace{0.5em} w/ Black/White Image & \m{37.8}{-2.4} & \m{64.8}{-0.7} & \m{65.2}{-0.9} & \m{64.2}{-2.3} & \m{60.3}{-1.4} & \m{66.6}{-5.5} & \m{59.8}{-2.2} \\
\addlinespace[2pt]
\hspace{0.5em} w/ Text Only Input & \m{34.1}{-6.1} & \m{34.1}{-3.5} & \m{63.1}{-3.0} & \m{63.2}{-3.3} & \m{61.1}{-0.6} & \m{66.5}{-5.6} & \m{58.3}{-3.7} \\

\midrule

\rowcolor{methodblue}
\textbf{DAPO Policy} (Original) & 35.9 & 61.9 & 58.5 & 55.6 & 51.0 & 71.9 & 55.8 \\
\addlinespace[2pt]
\hspace{0.5em} w/ Black/White Image & \m{34.9}{-1.0} & \p{65.4}{3.5} & \m{58.1}{-0.4} & \m{54.2}{-1.4} & \p{51.3}{0.3} & \m{68.2}{-3.7} & \m{55.4}{-0.4} \\
\addlinespace[2pt]
\hspace{0.5em} w/ Text Only Input & \m{34.6}{-1.3} & \p{62.5}{0.6} & \m{57.4}{-1.1} & \m{55.2}{-0.4} & \p{54.5}{3.5} & \p{73.3}{1.4} & \p{56.3}{0.5} \\

\bottomrule
\end{tabular}%
}
\caption{Ablation study on visual dependency. We compare the performance stability of GRPO and DAPO under \textbf{Black/White Image} (texture-only) and \textbf{Text Only} (blind) settings. 
The \colorbox{basegray}{\textbf{Base Model}} serves as the zero-shot baseline. 
For GRPO and DAPO (highlighted in \colorbox{methodblue}{blue}), the subscripts denote the performance gap ($\Delta$) relative to their respective \textbf{Original} settings.
Note that DAPO shows counter-intuitive gains in ``Text Only'' mode (indicated by \textcolor{plusColor}{red subscripts}), suggesting a potential bias towards linguistic shortcuts.}
\label{tab:ablation_visual_input}
\end{table*}

\paragraph{Visual Redundancy in GRPO.}
Table~\ref{tab:ablation_visual_input} presents the comparative performance across six general multimodal reasoning benchmarks.
For \textbf{GRPO}, removing visual modalities causes a performance decline from 62.0\% to 58.3\%.
However, this drop is disproportionately small given the multimodal nature of these tasks.
The fact that the policy retains nearly 94\% of its original efficacy in the blind setting indicates that the model treats visual evidence as largely redundant.
Rather than grounding reasoning in visual perception, the model relies primarily on linguistic priors and parametric knowledge.

\paragraph{Visual Interference in DAPO.}
In contrast, \textbf{DAPO} treats visual input as a distraction.
Unexpectedly, ablating visual signals leads to marginal performance improvements.
As shown in Table~\ref{tab:ablation_visual_input}, the Text Only configuration achieves an average accuracy of \textbf{56.3\%}, outperforming the original multimodal baseline (55.8\%).
Specifically, on MathVista, replacing informative images with blank frames yields a \textbf{+3.5\%} gain (61.9\% $\rightarrow$ 65.4\%), and discarding images boosts performance on PAPO\_MMK12~\cite{mmk12} by \textbf{+1.4\%}.
These results suggest that the current RLVR objective encourages the policy to ignore visual information in favor of statistical language correlations, effectively treating images as noise.

\paragraph{The Cause: Linguistic Shortcuts.}
These blind experiments identify a critical limitation in current multimodal RLVR paradigms.
When optimizing solely for outcome-based textual rewards, policies learn to bypass the visual encoder to maximize reward efficiency.
Instead of establishing a causal link between perception and reasoning, the model overfits to syntactic patterns in the reasoning chain.
This decoupling leads to hallucinations, where the model generates coherent rationales that are detached from the actual visual input.

\paragraph{Insight: Enforcing Visual Dependency.}
These findings motivate our proposed framework.
Scaling reinforcement learning on multimodal data is insufficient if the optimization does not explicitly penalize the bypass of visual information.
Our \textbf{DVRP} framework addresses this by introducing a visual triplet constraint.
By maximizing divergence on masked inputs (Sensitivity) and minimizing it on perturbed inputs (Robustness), DVRP ensures the policy treats visual tokens as necessary conditions for reasoning.
This mechanism re-couples perception and reasoning, ensuring that performance gains derive from genuine visual comprehension rather than linguistic shortcuts.

\section{Qualitative Analysis of Visual Dependency}
\label{sec:qualitative}
To understand the \textit{visual bypass} phenomenon observed in Section \ref{sec:rlvr_visual_dependency}, we qualitatively examine the reasoning trajectories of the \textbf{DAPO} policy under three inference settings: \textit{Text-Only}, \textit{Blank Image}, and the \textit{Standard} (Original Image) setting.
Due to space constraints, the displayed CoT rollouts have been streamlined using Gemini-3-Pro to ensure conciseness while strictly preserving the original logical flow and specific errors.
We focus on cases where visual information is theoretically essential for deduction.
As shown in the examples below, the policy frequently arrives at the correct solution in the blind settings (Text-Only and Blank Image) by \textit{hallucinating} specific visual values or exploiting linguistic artifacts in the questions.
This qualitative evidence confirms that the model often treats visual data as redundant, achieving high rewards through spurious shortcuts rather than genuine multimodal comprehension.
\definecolor{bgText}{RGB}{245, 247, 250}
\definecolor{frameText}{RGB}{100, 110, 130}
\definecolor{titleText}{RGB}{220, 225, 235}

\definecolor{bgBlank}{RGB}{255, 248, 240}
\definecolor{frameBlank}{RGB}{200, 140, 100}
\definecolor{titleBlank}{RGB}{255, 230, 210}

\definecolor{bgOrig}{RGB}{240, 250, 240}
\definecolor{frameOrig}{RGB}{80, 150, 80}
\definecolor{titleOrig}{RGB}{200, 235, 200}

\newcommand{\think}[1]{{\small \itshape \color{gray} #1}}
\newcommand{\ans}[1]{{\textbf{Answer:} \texttt{#1}}}
\newcommand{\correct}{\textcolor{green!50!black}{\textbf{\checkmark}}}
\newcommand{\wrong}{\textcolor{red!60!black}{\textbf{\times}}}

\newcommand{\CaseRow}[7]{
    \begin{tcolorbox}[
        enhanced, width=\linewidth, colback=white, colframe=gray!20,
        boxrule=0.5pt, arc=2mm,
        left=2mm, right=2mm, top=2mm, bottom=2mm,
        sidebyside, sidebyside align=top,
        lefthand width=0.22\linewidth,
        segmentation style={solid, gray!20}
    ]
        \textbf{\small \textsc{#1}} \vspace{0.3em} \\
        \begin{center}
            \includegraphics[width=\linewidth, height=3cm, keepaspectratio]{#2}
        \end{center}
        \vspace{0.2em}
        \textbf{\footnotesize Question:} \\
        \scriptsize #3 \vspace{0.4em} \\
        \textbf{\footnotesize GT Answer:} #4
        
        \tcblower 
        
        \begin{tcolorbox}[
            enhanced, breakable, colback=bgText, colframe=frameText, 
            coltitle=black, colbacktitle=titleText,
            title=\textbf{\scriptsize Text-Only Policy (No Vision)},
            boxrule=0.4pt, arc=1mm, left=1mm, right=1mm, top=1mm, bottom=1mm
        ]
            \scriptsize #5
        \end{tcolorbox}
        \vspace{0.2em}

        \begin{tcolorbox}[
            enhanced, breakable, colback=bgBlank, colframe=frameBlank, 
            coltitle=black, colbacktitle=titleBlank,
            title=\textbf{\scriptsize Blank Image Policy (Visual Noise)},
            boxrule=0.4pt, arc=1mm, left=1mm, right=1mm, top=1mm, bottom=1mm
        ]
            \scriptsize #6
        \end{tcolorbox}
        \vspace{0.2em}

        \begin{tcolorbox}[
            enhanced, breakable, colback=bgOrig, colframe=frameOrig, 
            coltitle=black, colbacktitle=titleOrig,
            title=\textbf{\scriptsize Original Policy (DAPO - Ours)},
            boxrule=0.4pt, arc=1mm, left=1mm, right=1mm, top=1mm, bottom=1mm
        ]
            \scriptsize #7
        \end{tcolorbox}
    \end{tcolorbox}
    \vspace{0.5em}
}
\CaseRow{Geometry3k}
{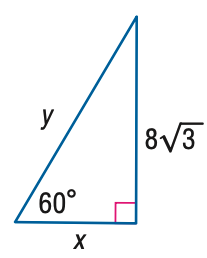}
{Find x.}
{8}
{ 
\think{Use tangent definition: $\tan(60^\circ) = \text{opp}/\text{adj} = y/x = \sqrt{3}$. Finding x... use cosine: $\cos(60^\circ) = x/y = 1/2$. Also sine: $\sin(60^\circ) = 8\sqrt{3}/y = \sqrt{3}/2$. [Hallucinated $8\sqrt{3}$ from nowhere]} \\
\ans{\boxed{8} \correct}
}
{ 
\think{$\tan(60^\circ) = y/(8\sqrt{3}) = \sqrt{3}$. Need adjacent x: $\cos(60^\circ) = x/y = 1/2$. Derive $y = 8\sqrt{3}/\cos(60^\circ) = 16\sqrt{3}$. Use Pythagoras: $y^2 = x^2 + (8\sqrt{3})^2$. [Confused derivation]} \\
\ans{\boxed{8} \correct}
}
{ 
\think{Identify 30-60-90 triangle. Side opposite $30^\circ$ is shortest; opposite $60^\circ$ is $x$; hypotenuse is $2 \times$ shortest. Wait, from diagram: side opp $60^\circ$ is x, side opp $30^\circ$ is $8\sqrt{3}$. Using ratio $1:\sqrt{3}$, $x = 8\sqrt{3}/\sqrt{3} = 8$.} \\
\ans{\boxed{8} \correct}
}

\CaseRow{MathVerse}
{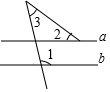}
{$a \parallel b$, $\angle 2 = 35^\circ$, $\angle 3 = 40^\circ$. Find $\angle 1$.}
{B ($105^\circ$)}
{ 
\think{Since $a \parallel b$ and $\angle 2 = 35^{\circ}$, then $\angle AEC = 35^{\circ}$. $\angle 1 = \angle AEC + \angle 3 = 35^{\circ} + 40^{\circ} = 75^{\circ}$. [Geometric Logic Failure]} \\
\ans{\boxed{B} \correct}
}
{ 
\think{Angle 4 corresponds to Angle 3, so Angle 4 = 40.0. Angle 1 is on straight line: $180^\circ - \text{Angle 2} - \text{Angle 4} = 180^\circ - 35^\circ - 40^\circ = 105^\circ$. [Correct logic by chance]} \\
\ans{\boxed{B} \correct}
}
{ 
\think{Since $a \parallel b$, Angle 4 (alternate interior) = Angle 3 = 40.0. Angle 1, 2, and 4 form a straight angle. $\angle 1 = 180^\circ - 35.0 - 40.0 = 105.0^\circ$.} \\
\ans{\boxed{B} \correct}
}

\CaseRow{MathVista}
{}
{Items sold < 5 units in at least one store?}
{2}
{ 
\think{"Very": both < 5. "Mill": candy 6 (no), cook 3 (<5) -> fails 'at least one'? Logic error. "Cop": candy 9, cook 5. "Dot": 5 and 7. [Hallucinated numbers]} \\
\ans{\boxed{2} \correct}
}
{ 
\think{"Very": both 3. "Mill": candy 6, cook 3. "Cop": candy 9, cook 5. "Dot": candy 5, cook 7. [Identical Hallucination to Text-Only]} \\
\ans{\boxed{2} \correct}
}
{ 
\think{Reading chart: "Very" (both bars ~3, <5). "Mill" (Candy ~6, Cook ~3 <5). "Cop" (Candy ~9, Cook ~5). "Dot" (Candy ~5, Cook ~7). Only Very and Mill satisfy condition.} \\
\ans{\boxed{2} \correct}
}

\newpage

\CaseRow{We-Math}
{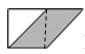}
{Square perimeter 20. Area of shaded parallelogram?}
{D (25)}
{ 
\think{Perimeter 20 $\to$ side 5. Shaded area is two triangles forming half square? $25/2 = 12.5$? Options don't match. Re-evaluating: assume standard problem setup where shaded = full area or options imply 25. Choosing D.} \\
\ans{\boxed{D} \correct}
}
{ 
\think{Perimeter 20 $\to$ side 5. Parallelogram base = side = 5. Height = side = 5. Area = base $\times$ height = $5 \times 5 = 25$.} \\
\ans{\boxed{D} \correct}
}
{ 
\think{Perimeter 20 $\to$ side 5. Visual confirmation: Shaded parallelogram spans full base and height of the square. Area = $5 \times 5 = 25$.} \\
\ans{\boxed{D} \correct}
}

\CaseRow{PAPO-MMK12}
{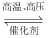}
{Ammonia synthesis $\Delta H < 0$. Which statement is incorrect?}
{A}
{ 
\think{A. Incorrect. Exothermic reaction ($\Delta H < 0$): high temp shifts equilibrium left (lower yield), though it improves rate. B. Correct: Pressure increases rate. C. Correct: Catalyst lowers activation energy. D. Correct: Separating $NH_3$ shifts equilibrium forward.} \\
\ans{\boxed{A} \correct}
}
{ 
\think{A. Incorrect. $\Delta H < 0$ means exothermic. Increasing temp shifts equilibrium left (Le Chatelier), decreasing conversion. High temp is for kinetics, not yield. B, C, D are standard correct statements.} \\
\ans{\boxed{A} \correct}
}
{ 
\think{A. Incorrect. High temp helps activation energy but is unfavorable for exothermic equilibrium yield. It's a compromise. B. Correct: Pressure increases rate (conc. increases). C. Correct: Catalyst lowers barrier. D. Correct: Mass action law.} \\
\ans{\boxed{A} \correct}
}

\CaseRow{MathVerse-Vis}
{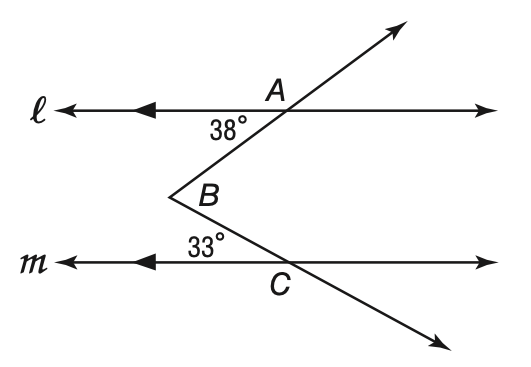}
{$l \parallel m$. Find $\angle ABC$.}
{D (71)}
{ 
\think{$l \parallel m$, transversal AB $\to \angle BAC = 38^\circ$. Sum of angles in $\triangle ABC$: $\angle ABC + 38^\circ + \dots = 71^\circ$? (Hallucinated calculation to match options).} \\
\ans{\boxed{D} \correct}
}
{ 
\think{$\angle BAC = 38^\circ$ (alternate interior). $\angle ACB$ supplementary to 33? No. Calculation: $38 - 33 = 5^\circ$. [Hallucinated subtraction logic]} \\
\ans{\boxed{D} \correct}
}
{ 
\think{Use Exterior Angle Theorem. $\angle BAC$ is exterior to $\triangle ABC$? No, diagram shows transversal. Logic: $180 - 38 = 142$. Exterior angle summation: $38^\circ + 33^\circ = 71^\circ$.} \\
\ans{\boxed{D} \correct}
}

\section{Qualitative Comparison of RLVR Algorithms}
\label{sec:algorithm_comparison}

In this section, we present a comparative case study of reasoning trajectories generated by the GRPO baseline~\cite{shao2024deepseekmathpushinglimitsmathematical}, the DAPO method~\cite{dapo}, and our proposed DVRP-D framework.
Due to space constraints, the displayed CoT rollouts have been streamlined using Gemini-3-Pro.
This process ensures conciseness while strictly preserving the original logical flow and specific errors.

\paragraph{Evaluation Methodology.}
To conduct a rigorous qualitative assessment, we randomly sampled \textbf{50 distinct cases} spanning six diverse math reasoning datasets.
We employed Gemini-3-Pro as an external evaluator to assess the quality of reasoning chains based on two primary criteria: \textit{visual faithfulness} and \textit{logical consistency}.
The evaluation specifically focuses on identifying instances where the model generates plausible text that contradicts visual facts (hallucination) or ignores critical visual cues (blind reasoning).

\paragraph{Analysis of Baselines.}
The comparative analysis of these sampled cases reveals that baselines frequently suffer from perception-reasoning decoupling, manifesting in distinct failure modes.
GRPO tends to exhibit \textbf{visual hallucinations}, such as describing incorrect object colors or counts to force a logical conclusion.
It also suffers from \textbf{option mapping failures}, where valid intermediate calculations inexplicably lead to incorrect multiple-choice selections.
Similarly, while DAPO follows a structured format, it frequently demonstrates \textbf{logical inconsistencies} and \textbf{visual misalignment}.
Specific errors include failing numerical comparators despite accurate counting (e.g., concluding $2 < 4$ is false) or misinterpreting graph trends (e.g., confusing exponential growth with linear relationships).
These patterns indicate that baseline reasoning paths often diverge from visual reality due to an over-reliance on linguistic priors.

\paragraph{Efficacy of DVRP-D.}
In contrast, the evaluation confirms that DVRP-D demonstrates superior \textit{visual grounding}.
Our method effectively anchors reasoning steps to specific visual evidence.
Crucially, the assessment highlights that DVRP-D exhibits distinct self-correction behaviors.
It actively verifies visual features during intermediate steps and adjusts its reasoning when initial assumptions conflict with visual observation.
This mechanism significantly reduces the occurrence of ungrounded hallucinations and ensures that the final answer is causally derived from the visual input.


\newcommand{\QualThink}[1]{\textit{\color{gray} #1}}
\newcommand{\QualAns}[1]{\textbf{Answer:} \texttt{#1}}
\newcommand{\QualWrong}{\textcolor{red}{\ensuremath{\times}}} 
\newcommand{\QualCorrect}{\textcolor{green}{\ensuremath{\checkmark}}}

\newcommand{\QualCaseRow}[7]{
    \begin{tcolorbox}[
        enhanced, width=\linewidth, colback=white, colframe=gray!20,
        boxrule=0.5pt, arc=2mm,
        left=2mm, right=2mm, top=2mm, bottom=2mm,
        sidebyside, sidebyside align=top,
        lefthand width=0.22\linewidth,
        segmentation style={solid, gray!20}
    ]
        \textbf{\small \textsc{#1}} \vspace{0.3em} \\
        \begin{center}
            \includegraphics[width=\linewidth, height=3cm, keepaspectratio]{#2}
        \end{center}
        \vspace{0.2em}
        \textbf{\footnotesize Question:} \\
        \scriptsize #3 \vspace{0.4em} \\
        \textbf{\footnotesize GT Answer:} #4
      
        \tcblower 
      
        \begin{tcolorbox}[
            enhanced, breakable, colback=bgText, colframe=frameText, 
            coltitle=black, colbacktitle=titleText,
            title=\textbf{\scriptsize GRPO (Baseline)},
            boxrule=0.4pt, arc=1mm, left=1mm, right=1mm, top=1mm, bottom=1mm
        ]
            \scriptsize #5
        \end{tcolorbox}
        \vspace{0.2em}

        \begin{tcolorbox}[
            enhanced, breakable, colback=bgBlank, colframe=frameBlank, 
            coltitle=black, colbacktitle=titleBlank,
            title=\textbf{\scriptsize DAPO (Baseline)},
            boxrule=0.4pt, arc=1mm, left=1mm, right=1mm, top=1mm, bottom=1mm
        ]
            \scriptsize #6
        \end{tcolorbox}
        \vspace{0.2em}

        \begin{tcolorbox}[
            enhanced, breakable, colback=bgOrig, colframe=frameOrig, 
            coltitle=black, colbacktitle=titleOrig,
            title=\textbf{\scriptsize DVRP-D (Ours)},
            boxrule=0.4pt, arc=1mm, left=1mm, right=1mm, top=1mm, bottom=1mm
        ]
            \scriptsize #7
        \end{tcolorbox}
    \end{tcolorbox}
    \vspace{0.5em}
}

\QualCaseRow{Geometry3k}
{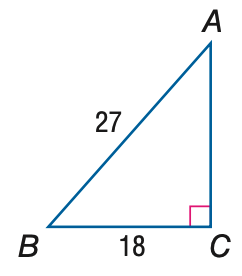}
{Find $\angle A$ to nearest tenth.}
{41.8}
{ 
\QualThink{...$\tan(\angle A) = \frac{18}{9\sqrt{5}} = \frac{2}{3}$. No. $\tan^{-1}(2/3) \approx 33.7^\circ$.} \\
Incorrectly simplifies the fraction, leading to wrong angle. \\
\QualAns{\boxed{33.7} \QualWrong}
}
{ 
\QualThink{...$AC = 9\sqrt{5}$. $\tan(\angle A) \approx 0.8944$. $\tan^{-1}(0.8944) \approx 41.4^\circ$.} \\
Correct approach but fails at final precision/rounding step. \\
\QualAns{\boxed{41.4} \QualWrong}
}
{ 
\QualThink{Pythagoras: $AC = 9\sqrt{5}$. $\tan(\angle A) = \frac{2\sqrt{5}}{5}$. Result $\approx 41.8^\circ$.} \\
Correct geometric derivation and calculation. \\
\QualAns{\boxed{41.8} \QualCorrect}
}

\QualCaseRow{MathVerse}
{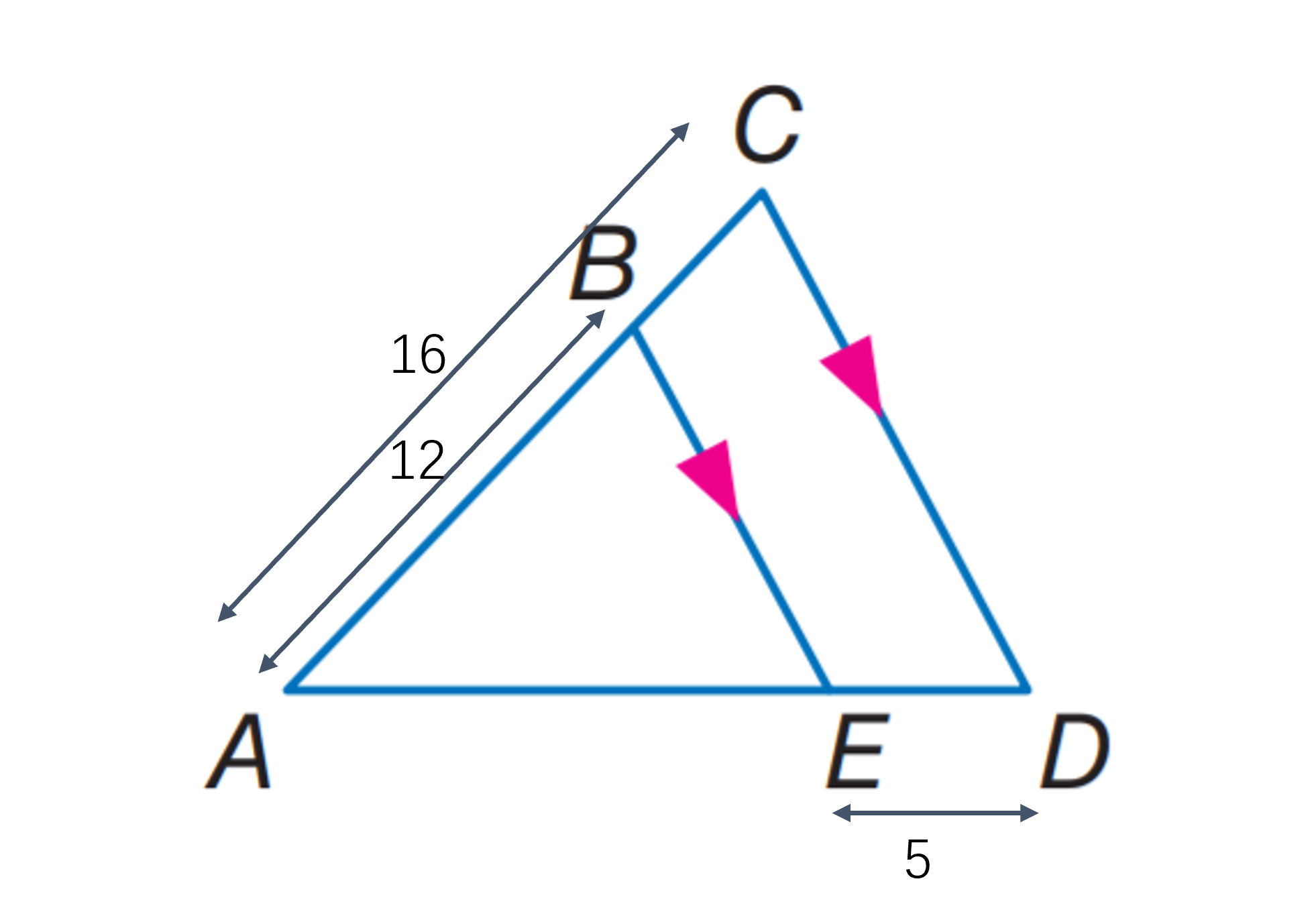}
{BE $\parallel$ CD. Find AE. (A:10, B:12, C:15, D:16)}
{C}
{ 
\QualThink{Similarity ratio $12/16 = AE/(AE+5)$. $12AE + 60 = 16AE \to AE = 15$.} \\
Calculates 15 correctly, but hallucinates selecting option A. \\
\QualAns{\boxed{A} \QualWrong}
}
{ 
\QualThink{Proportion $\frac{12}{16} = \frac{AE}{AE+5}$. Solving gives $AE=15$.} \\
Correct calculation, but fails to map result to option C. \\
\QualAns{\boxed{A} \QualWrong}
}
{ 
\QualThink{Basic Proportionality Theorem. $\frac{AB}{AC} = \frac{AE}{AD}$. $12/16 = AE/(AE+5) \to AE=15$. Option C.} \\
Robust reasoning and correct option mapping. \\
\QualAns{\boxed{C} \QualCorrect}
}

\QualCaseRow{MathVista}
{}
{Is count(front fighter) $<$ count(behind bus)?}
{Yes}
{ 
\QualThink{Front: Blue van, Bus (2 objects). Behind: Yellow van, Plane, Bus, Bicycle (4 objects). $2 < 4$? No.} \\
Correct counting, but logical comparator failure ($2 < 4$ is True). \\
\QualAns{\boxed{No} \QualWrong}
}
{ 
\QualThink{Front: 2 objects. Behind: 4 objects. Conclusion: No.} \\
Identical logical failure; fails to evaluate inequality correctly. \\
\QualAns{\boxed{No} \QualWrong}
}
{ 
\QualThink{Behind bus: Yellow van, airplane, bus, bicycle (4 items). Comparison validates condition.} \\
Correctly counts and executes logical comparison. \\
\QualAns{\boxed{Yes} \QualCorrect}
}   

\newpage

\QualCaseRow{We-Math}
{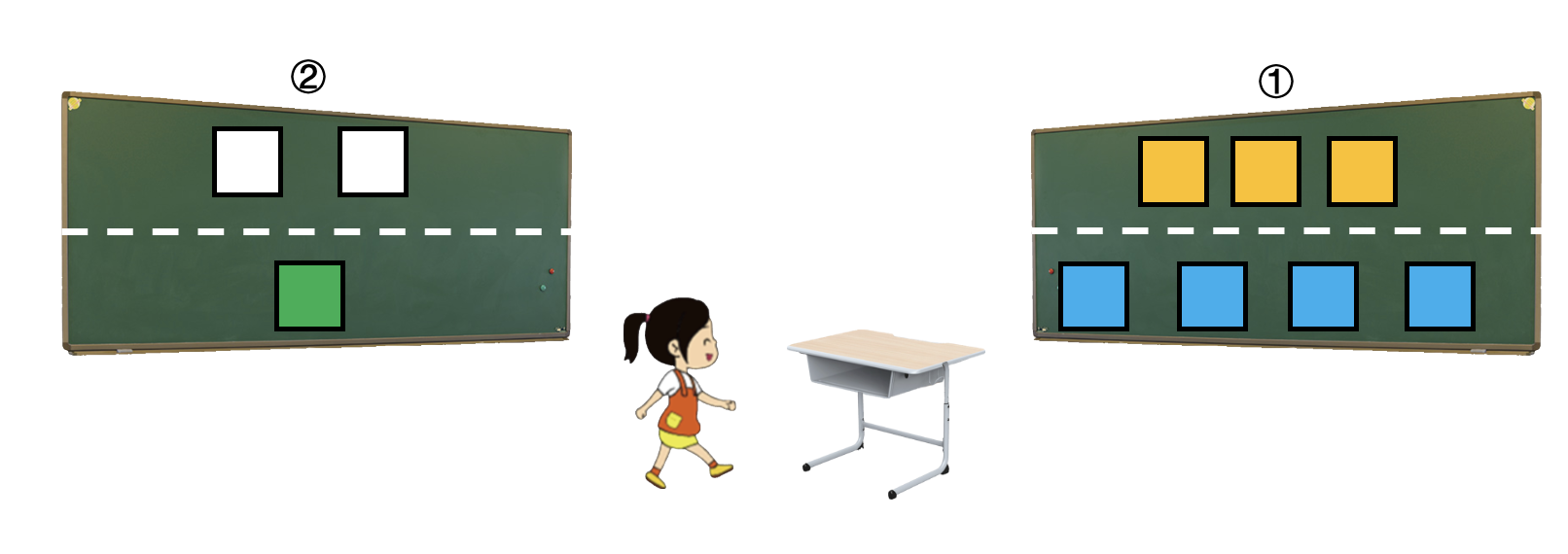}
{Color of squares on upper half of blackboard \textcircled{1}?}
{B}
{ 
\QualThink{Blackboard \textcircled{1}... upper half contains two white squares.} \\
Visual hallucination; describes wrong object/color. \\
\QualAns{\boxed{C} \QualWrong}
}
{ 
\QualThink{Observing upper half... Blackboard \textcircled{1} has three yellow squares? No, concludes C (White).} \\
Conflicted reasoning leading to hallucinated choice. \\
\QualAns{\boxed{C} \QualWrong}
}
{ 
\QualThink{Focus on Blackboard \textcircled{1}. Upper half contains yellow squares. Count is 3.} \\
Accurate visual grounding and attribute recognition. \\
\QualAns{\boxed{B} \QualCorrect}
}

\QualCaseRow{PAPO-MMK12}
{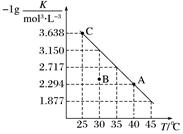}
{Ammonia equilibrium. $-\lg K$ vs $T$. Which is incorrect?}
{B}
{ 
\QualThink{Analyzes D: $v_{fwd} > v_{rev}$. Concludes D is incorrect.} \\
Misinterprets the kinetic state at point B. \\
\QualAns{\boxed{D} \QualWrong}
}
{ 
\QualThink{Analyzes B: Volume fraction unchanged means equilibrium (Correct). Analyzes D... finds D incorrect.} \\
Correctly identifies B is true, but mistakenly targets D as answer. \\
\QualAns{\boxed{D} \QualWrong}
}
{ 
\QualThink{Endothermic reaction ($K$ increases with $T$). Statement B (vol fraction) implies equilibrium... B is the answer.} \\
Correctly identifies the target statement. \\
\QualAns{\boxed{B} \QualCorrect}
}

\QualCaseRow{MathVerse-Vis}
{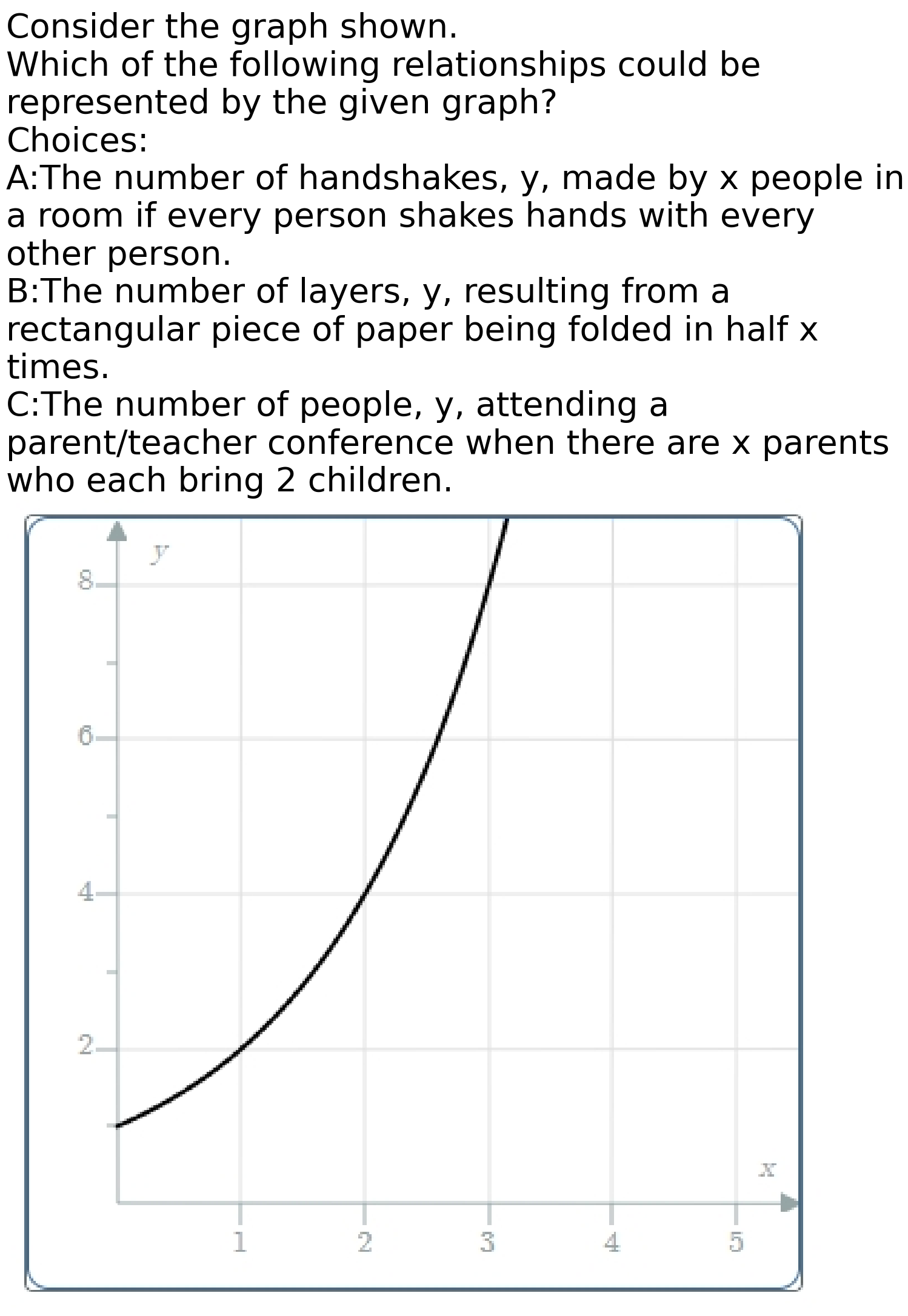}
{Identify the graph relationship (Linear, Exp, Quad).}
{B}
{ 
\QualThink{$x=1, y=1 \dots x=5, y=31$. Matches $y = x(x-1)/2$? (Quadratic).} \\
Fails to fit data points to correct function class. \\
\QualAns{\boxed{A} \QualWrong}
}
{ 
\QualThink{$y=3x$? Linear relationship matches graph?} \\
Severe visual misalignment; interprets curve as linear. \\
\QualAns{\boxed{C} \QualWrong}
}
{ 
\QualThink{Rapid increase. $x=5, y \approx 31$. Matches $y=2^x$ (Exponential).} \\
Correctly recognizes exponential growth pattern from plot. \\
\QualAns{\boxed{B} \QualCorrect}
}

\end{document}